 \documentclass[letterpaper]{article}
\usepackage{uai2019}
\usepackage[margin=1in]{geometry}

\usepackage{times}

\usepackage[utf8]{inputenc} 
\usepackage[T1]{fontenc}    
\usepackage{hyperref}       
\usepackage{url}            
\usepackage{booktabs}       
\usepackage{amsfonts}       
\usepackage{nicefrac}       
\usepackage{microtype}      

\usepackage{mathtools}
\usepackage{amsmath}
\usepackage{amssymb}

\usepackage{amsthm}

\usepackage{multirow}
\usepackage{url}

\usepackage{breakurl}
\usepackage{appendix}
\usepackage{makecell}
\usepackage{booktabs}

\usepackage[percent]{overpic}

\usepackage{bbm}

\usepackage{natbib}

\usepackage{wrapfig}

\usepackage{placeins}

\usepackage{makeidx}
\makeindex
\printindex

\title{Expressive Priors in Bayesian Neural Networks: \\ Kernel Combinations and Periodic Functions}

\author{
  Tim Pearce$^{1*}$, Russell Tsuchida$^2$, Mohamed Zaki$^1$, Alexandra Brintrup$^1$, Andy Neely$^1$ \\
  $^1$ Department of Engineering, University of Cambridge\\
  $^2$ School of ITEE, University of Queensland\\ 
  $^*$ \texttt{tp424@cam.ac.uk} \\
}

\usepackage{makeidx}
\makeindex
\printindex

\begin{document}

\maketitle

\begin{abstract}

A simple, flexible approach to creating expressive priors in Gaussian process (GP) models makes new kernels from a combination of basic kernels, e.g. summing a periodic and linear kernel can capture seasonal variation with a long term trend. Despite a well-studied link between GPs and Bayesian neural networks (BNNs), the BNN analogue of this has not yet been explored. This paper derives BNN architectures mirroring such kernel combinations. Furthermore, it shows how BNNs can produce periodic kernels, which are often useful in this context. These ideas provide a principled approach to designing BNNs that incorporate prior knowledge about a function. We showcase the practical value of these ideas with illustrative experiments in supervised and reinforcement learning settings. \footnote{Code for plots and experiments is available at: \newline \url{https://github.com/TeaPearce}} 


\end{abstract}

\section{INTRODUCTION}

\newcommand\widthPriors{0.24\textwidth}
\newcommand\PriorsTextHeight{8} 
\begin{figure}[t!]
\begin{center}
	 \vskip 0.0in
	 
	Basic BNNs
     \vskip 0.05in
     
    \begin{minipage}{\widthPriors}
    	\centering
        \begin{overpic}[width=\linewidth]{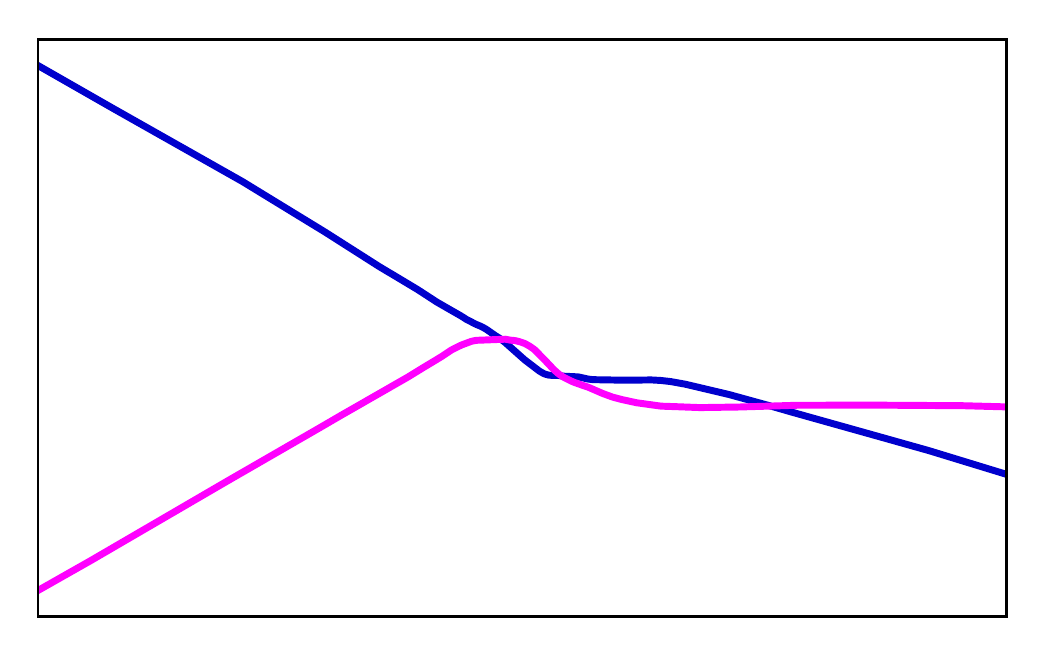}
         \put(50,\PriorsTextHeight){\makebox(0,0){ \small{ReLU}}}
        \end{overpic}
    \end{minipage} \hskip -0.05in
    \begin{minipage}{\widthPriors}
    	\centering
        \begin{overpic}[width=\linewidth]{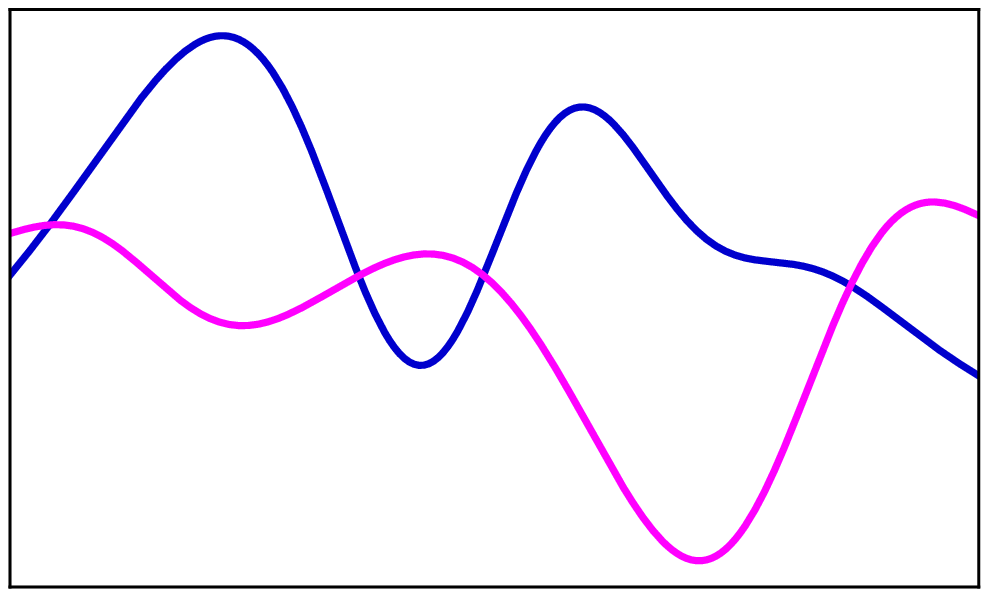}
        \put(50,\PriorsTextHeight){\makebox(0,0){ \small{RBF}}}
        \end{overpic}
    \end{minipage}
    
    \begin{minipage}{\widthPriors}
    	\centering
        \begin{overpic}[width=\linewidth]{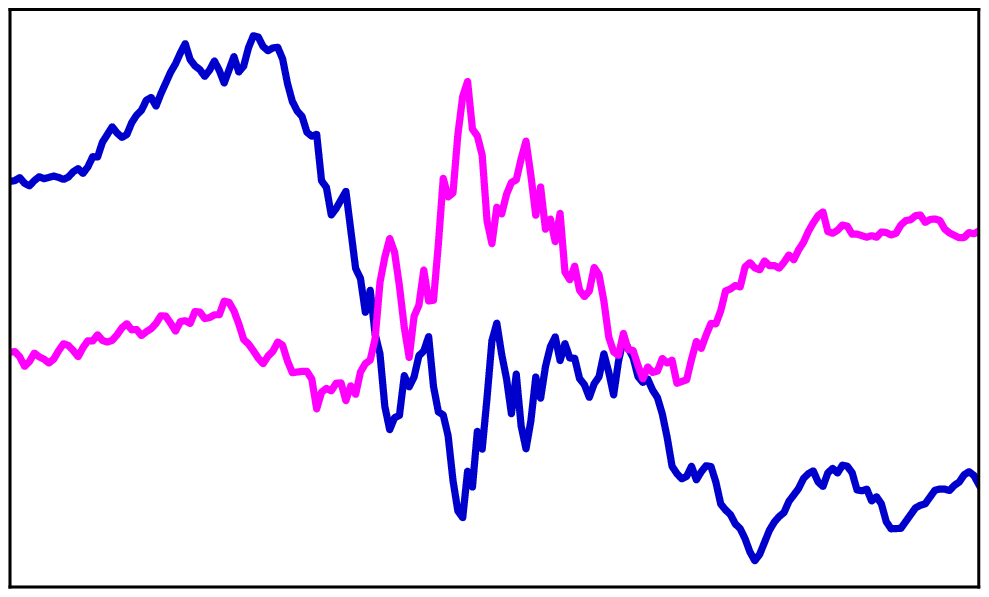}
         \put(50,\PriorsTextHeight){\makebox(0,0){ \small{ERF}}}
        \end{overpic}
    \end{minipage} \hskip -0.05in
    \begin{minipage}{\widthPriors}
    	\centering
        \begin{overpic}[width=\linewidth]{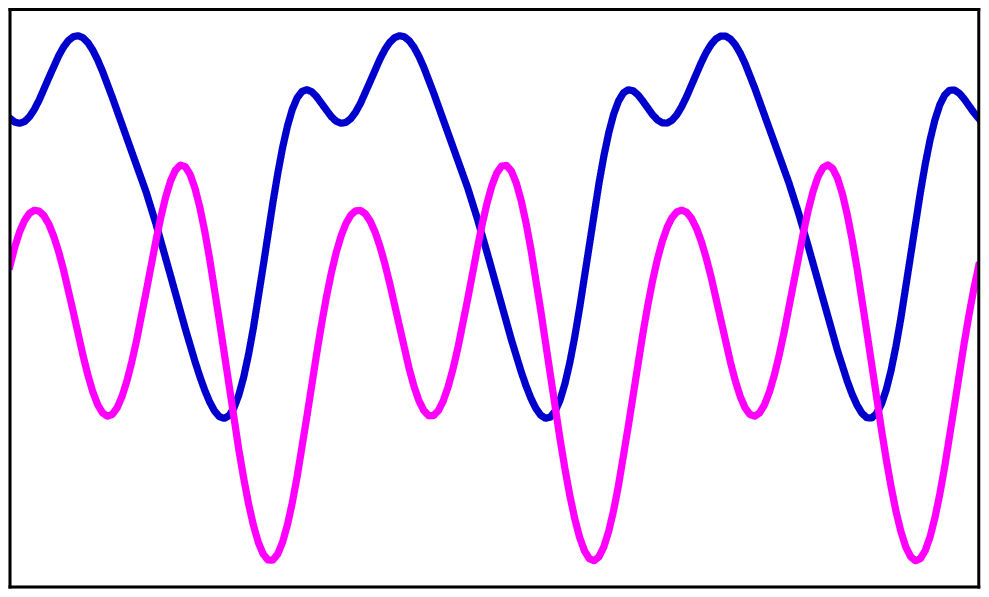}
        \put(50,\PriorsTextHeight){\makebox(0,0){ \small{Periodic}}}
        \end{overpic}
    \end{minipage}
    
    \vskip 0.15in
    Combinations of Basic BNNs
    \vskip 0.05in
    
    
    \begin{minipage}{\widthPriors}
    	\centering
        \begin{overpic}[width=\linewidth]{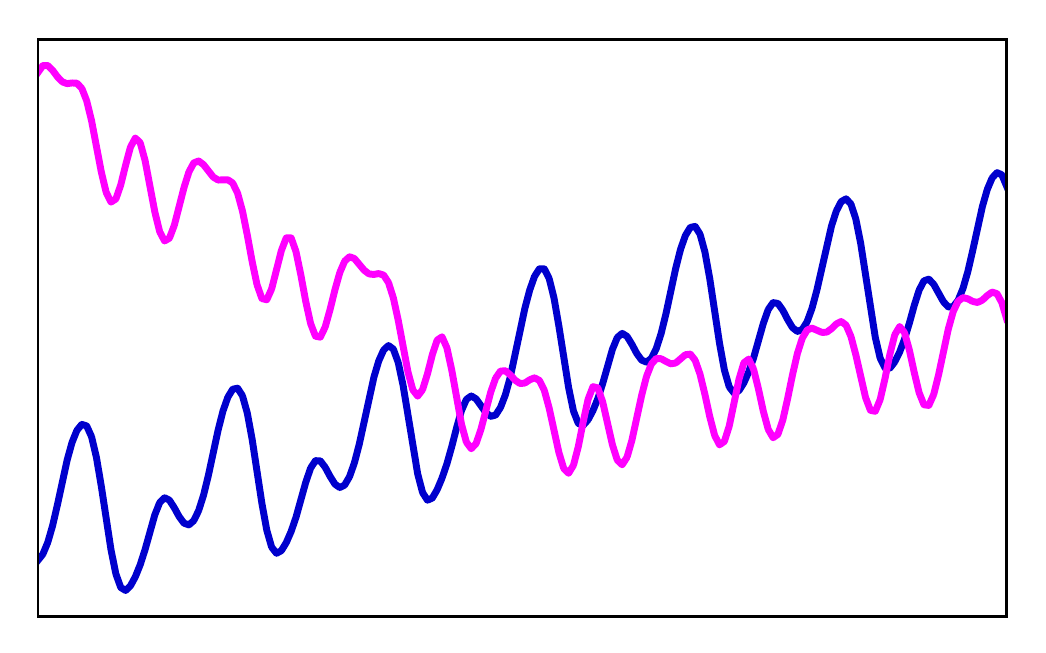}
         \put(50,\PriorsTextHeight){\makebox(0,0){ \small{Leaky ReLU $+$ Periodic}}}
        \end{overpic}
    \end{minipage} \hskip -0.05in
    \begin{minipage}{\widthPriors}
    	\centering
        \begin{overpic}[width=\linewidth]{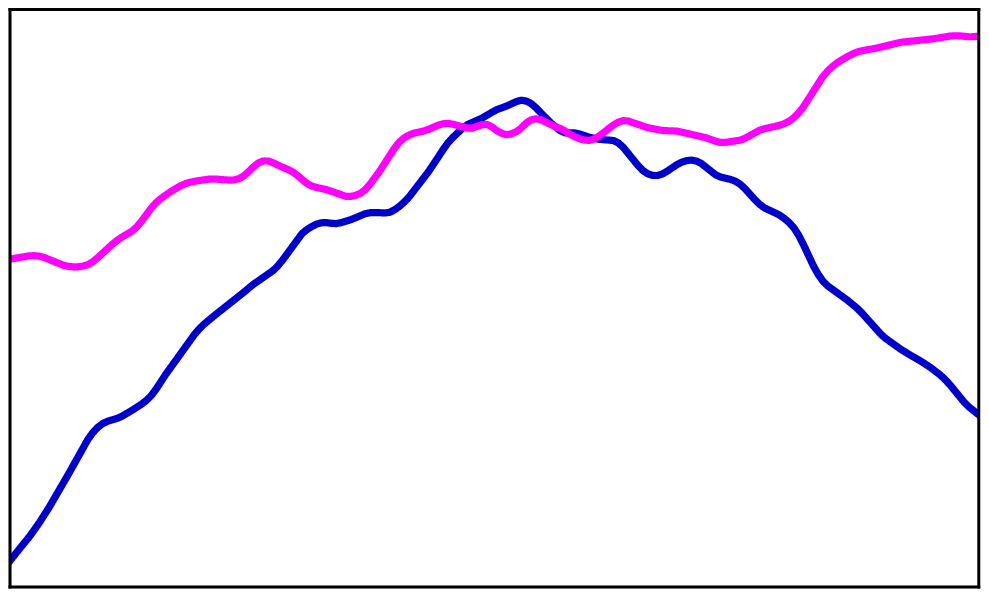}
        \put(50,\PriorsTextHeight){\makebox(0,0){ \small{ReLU $+$ ERF}}}
        \end{overpic}
    \end{minipage}

     
    \begin{minipage}{\widthPriors}
    	\centering
        \begin{overpic}[width=\linewidth]{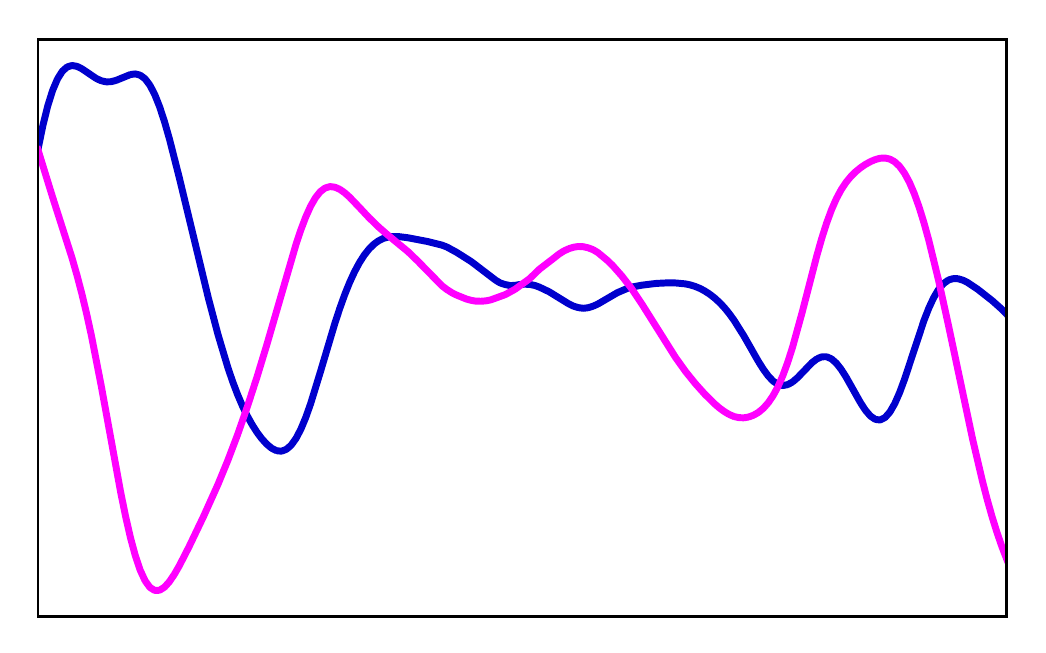}
         \put(50,\PriorsTextHeight){\makebox(0,0){ \small{ReLU $\times$ Periodic}}}
        \end{overpic}
    \end{minipage} \hskip -0.05in
    \begin{minipage}{\widthPriors}
    	\centering
        \begin{overpic}[width=\linewidth]{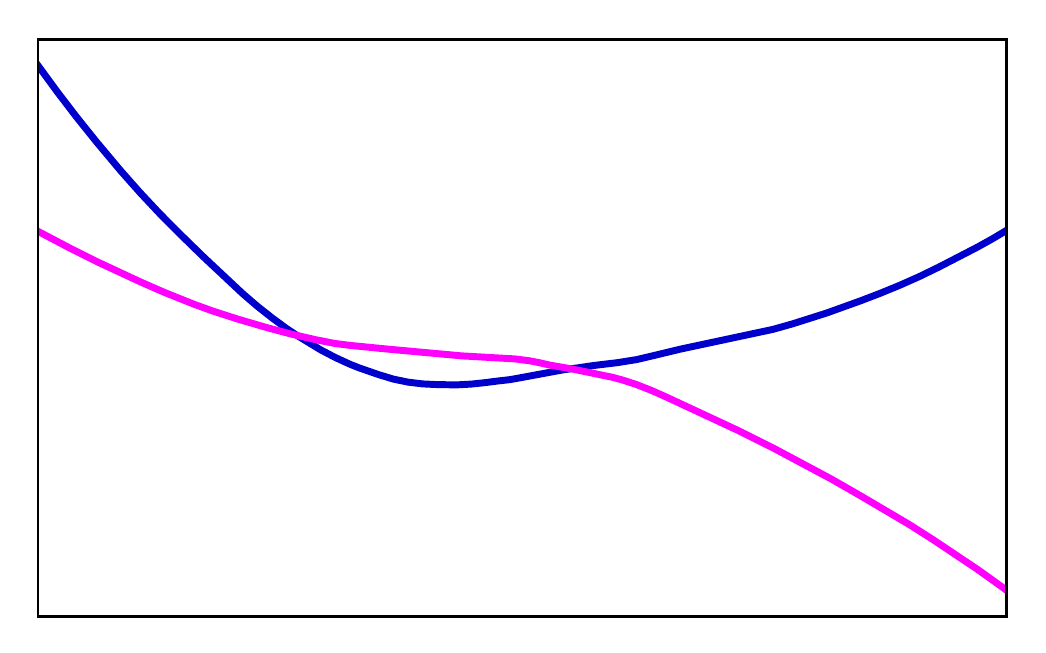}
        \put(50,\PriorsTextHeight){\makebox(0,0){ \small{ReLU $\times$ ReLU}}}
        \end{overpic}
    \end{minipage}
    
     
    \begin{minipage}{\widthPriors}
    	\centering
        \begin{overpic}[width=\linewidth]{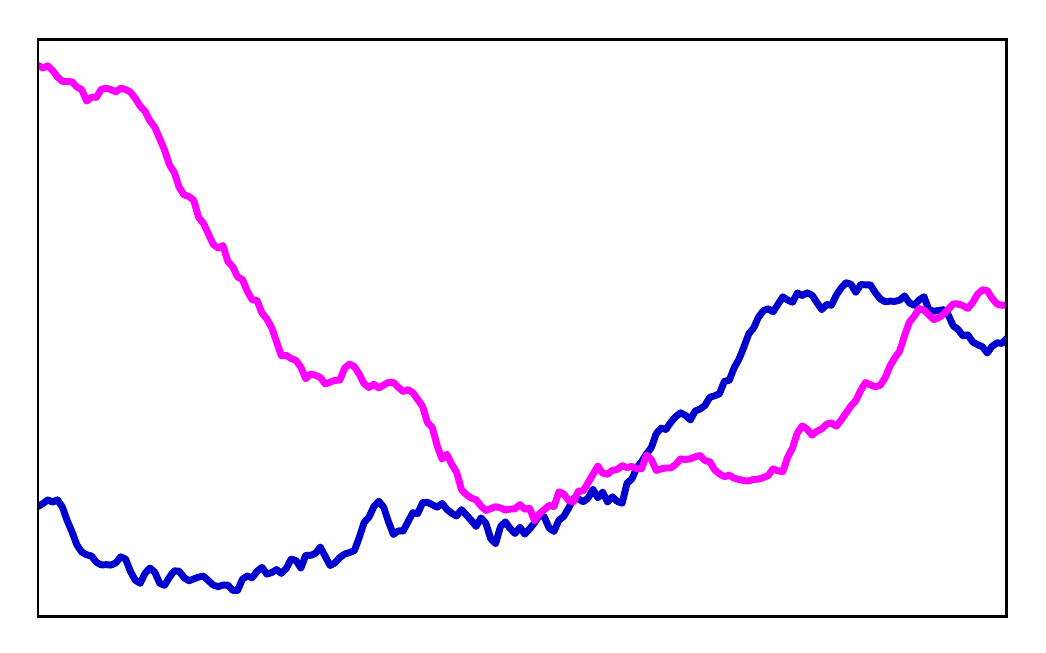}
         \put(50,\PriorsTextHeight){\makebox(0,0){ \small{$($ERF $+$ ERF$)$ $\times$ ReLU}}}
        \end{overpic}
    \end{minipage} \hskip -0.05in
    \begin{minipage}{\widthPriors}
    	\centering
        \begin{overpic}[width=\linewidth]{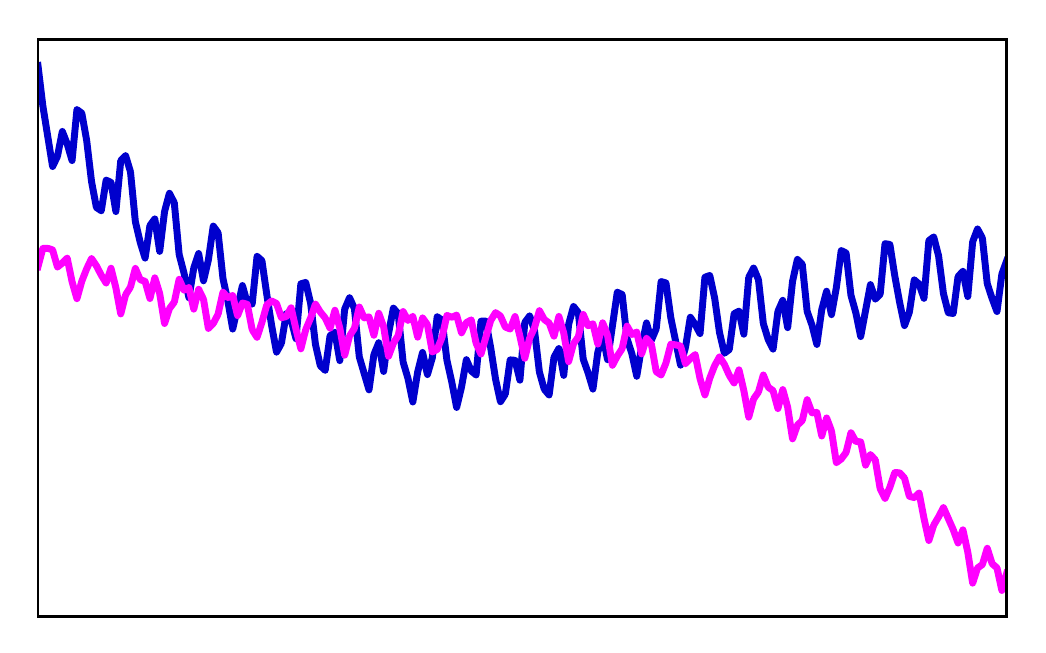}
        \put(50,\PriorsTextHeight){\makebox(0,0){ \small{$($ReLU $\times$ ReLU$)$ $+$ Periodic}}}
        \end{overpic}
    \end{minipage}
\caption{BNN architecture determines our prior belief about a function's properties. In general BNNs provide little flexibility in this regard - modifying only the activation function and length scale (`Basic BNNs'). This paper explores how to design BNNs to produce more expressive prior functions (`Combinations of Basic BNNs'). Two prior draws are shown for each BNN architecture.}
\label{fig_priors}
\end{center}
\vskip -0.4in
\end{figure}


One of deep learning's major achievements was mastering Atari games to human level, with each of the 49 games learnt using an identical algorithm, neural network (NN) architecture, and hyperparameters \citep{Mnih2015}.

Conversely, Gaussian process (GP) modelling places great emphasis on tailoring structure and hyperparameters to individual tasks - four pages of the seminal GP text are dedicated to the careful design of a kernel for a dataset of just 545 datapoints \citep{Rasmussen2018} [p118-122]. Indeed this incorporation of relevant prior knowledge is central to all Bayesian methods.

Bayesian neural networks (BNNs) lie at the curious intersection between these two modelling philosophies. They have strong theoretical links to GPs \citep{Neal1997}, yet ultimately share architectures with deep learning models.

The majority of previous research on BNNs has focused on developing methods for efficient inference \citep{Neal1997, Hernandez-Lobato2015, Blundell2015}, and, more recently, how they can benefit learning frameworks \citep{Gal2017a, Nguyen2018}. 


Relatively little work has explored prior design in BNNs - current wisdom takes an architecture expected to work well in non-Bayesian NNs, and places distributions over the weights. This can ignore significant prior information  humans bring to tasks \citep{Dubey2018}. The observation that \textbf{there seems little point in adopting a Bayesian framework if we don't, and can't, specify effective priors}, forms the core motivation for this paper. 

It is well known that BNNs converge to GPs \citep{Neal1997}. Whilst correspondence is only exact for BNNs of infinite width, this provides a useful lens through which to study the relationship between BNN architecture and prior.


The paper begins with an overview of this connection, discussing priors over functions produced by basic BNNs (which we define as fully-connected feed-forward NNs with iid Gaussian priors over parameters), and the effect of their hyperparameters. Our major contribution then follows in section \ref{sec_kernel_combos_BNN}: we consider porting an idea for prior design in GPs to BNNs. A simple approach to building expressive priors in GPs is to combine basic kernels to form a new kernel. We derive BNN architectures mirroring these effects. Figure \ref{fig_priors} shows examples of priors that can be expressed by basic BNNs, followed by the richer class of priors that can be expressed using these combined BNN architectures.

One situation where kernel combinations are useful is for functions with imperfect periodicity. This property is easily captured by combining a periodic kernel with some other kernel. We explore periodicity in BNNs in section \ref{sec_periodic_BNNs}, showing it is not enough to simply use cosine activations, as might be expected. We develop an alternative approach that precisely recovers a popular GP periodic kernel.


Illustrative experiments in section \ref{sec_experiments} showcase the practical value of our theoretical results both in supervised time series prediction and in reinforcement learning (RL) on a classic control task.

This paper is important from three perspectives. 
\begin{enumerate}
\item As a theoretical result further linking GPs with BNNs.
\item As a practical approach to creating more expressive priors in Bayesian deep learning models.
\item For non-Bayesian deep learning, enabling proper model specification for periodic and locally periodic functions.
\end{enumerate}

\section{BACKGROUND}
\label{sec_background}

\subsection{GAUSSIAN PROCESSES}

A GP is a stochastic process, fully described by its mean function, $\mathbb{E} [f(\mathbf{x})]$, and covariance function (or `kernel'), $K(\mathbf{x}, \mathbf{x}')$. Any finite subset of a GP's realisations follows a multivariate Gaussian distribution, which makes many analytical computations possible. They are considered a Bayesian non-parametic model in machine learning - see \cite{Rasmussen2018} for a full introduction. \cite{Duvenaud2014} provides a reference for the below.

A GP's mean function is often assumed zero, as it will be throughout this work. The kernel then determines the generalisation properties of the model. Informally, a kernel is a function that describes how closely two arbitrary data points, $\mathbf{x}$ \& $\mathbf{x}'$, are related. One might expect that if data points are similar, their outputs should also be similar. A common choice of kernel, squared exponential (SE), captures such behaviour,
\begin{equation}
\label{eq_SE_GP}
K_{SE}(x, x') = \sigma^2 \exp\bigg(-\frac{\lvert \lvert \mathbf{x} - \mathbf{x}' \rvert \rvert_2^2}{l^2} \bigg),
\end{equation}
where, $l$, the length scale, provides some control over how quickly similarity fades, and $\sigma^2$ is a scaling parameter.

The behaviour implied by the SE kernel is not suitable for all datasets. Data generated by a periodic function, for instance, would not follow this simple similarity rule. Here a periodic kernel is appropriate, e.g. the exponential sine squared kernel (ESS), for the 1-D case is, 
\begin{equation}
K_{ESS}(x,x') = \sigma^2 \exp \bigg( -  \frac{2 \sin^2(\frac{ \pi }{p} (x-x') ) }{l^2}   \bigg)
\label{eq_GPperiodic}
\end{equation}
where, $p$, determines the period over which the function repeats.

Many kernels are available, and selection of one that encodes properties of the function being modelled can be critical for good performance. By choosing kernels well suited to a problem, we are specifying appropriate priors.

What if a dataset has properties not well described by any of these kernels? A simple solution is to combine basic kernels together to make a new kernel. One can be surprisingly flexible in how this is done - directly multiplying or adding kernels, or applying warping to inputs \citep{Brito2003} [4.1]. This vastly increases the expressiveness of possible priors. 

Using the kernels from above for illustration, in order to model the function, $f(x)=\sin(x) + x$, one might choose, $K = K_{ESS} + K_{SE}$. For the function, $f(x)=x \sin(x)$, a good choice might be, $K = K_{ESS} \times K_{SE}$. In section \ref{sec_experiments} we model two time series with similar properties.


\subsection{BNNs CONVERGE TO GPs}
\label{sec_one_act}

Here we reproduce the derivation of infinitely wide single-layer BNNs as GPs \citep{Williams1996, Neal1997}.

Consider a single-layer NN, $f(\mathbf{x}): \mathbb{R}^d \to \mathbb{R}$, with input, $\mathbf{x}$, weights, $\mathbf{w}_1$ \& $\mathbf{w}_2$, biases $\mathbf{b}_1$, activation function, $\psi$, and hidden units $H$, with no final bias (to unclutter analysis),
\begin{equation}
\label{eq_NN}
f(\mathbf{x}) = \sum_{i=1}^H w_{2i} \psi( \mathbf{w}_{1i}\mathbf{x} + b_{1i} ).
\end{equation}
If priors centered at zero are placed over the parameters, we have a BNN with, $\mathbb{E}[f(\mathbf{x})] = 0$, hence the mean function is zero.
Consider now the covariance of outputs corresponding to two arbitrary inputs, $\mathbf{x}$ \& $\mathbf{x}'$. Denoting for convenience $\psi_i(\mathbf{x}) := \psi( \mathbf{w}_{1i}\mathbf{x} + b_{1i} )$,
\begin{equation}
\label{eq_}
K(\mathbf{x}, \mathbf{x}') = \mathbb{E}[f(\mathbf{x}) f(\mathbf{x}')]
\end{equation}
\begin{equation}
\label{eq_sum_cov}
 =  \mathbb{E} \bigg[  \big( \sum_{i=1}^H w_{2i} \psi_i(\mathbf{x})  \big)
   \big(   \sum_{j=1}^H w_{2j} \psi_j(\mathbf{x}')   \big)   \bigg]
\end{equation}
\begin{equation}
\label{eq_}
\begin{split}
 =  \mathbb{E} \big[   	w_{2,1}\psi_1(\mathbf{x}) w_{2,1}\psi_1(\mathbf{x}') +   w_{2,1}\psi_1(\mathbf{x}) w_{2,2}\psi_2(\mathbf{x}')  + \cdots \\
   w_{2,2}\psi_2(\mathbf{x}) w_{2,1}\psi_1(\mathbf{x}') +   w_{2,2}\psi_2(\mathbf{x}) w_{2,1}\psi_1(\mathbf{x}' )  + \cdots \\
 \cdots + w_{2,H}\psi_H(\mathbf{x}) w_{2,H}\psi_H(\mathbf{x}')
   \big]
\end{split}
\end{equation}
if parameter priors are independent, we find the terms between separate hidden units are zero, e.g. $\mathbb{E}[ w_{2,1}\psi_1(\mathbf{x}) w_{2,2}\psi_2(\mathbf{x}')]=\mathbb{E}[ w_{2,1} ] \mathbb{E}[\psi_1(\mathbf{x})] \mathbb{E}[  w_{2,2}] \mathbb{E}[\psi_2(\mathbf{x}')]=0$, so,
\begin{equation}
\label{eq_}
\begin{split}
 =  \mathbb{E} \big[   	w_{2,1}\psi_1(\mathbf{x}) w_{2,1}\psi_1(\mathbf{x}') +   w_{2,2}\psi_2(\mathbf{x}) w_{2,2}\psi_2(\mathbf{x}')  +  \cdots \\
 \cdots + w_{2,H}\psi_H(\mathbf{x}) w_{2,H}\psi_H(\mathbf{x}')
   \big]
\end{split}
\end{equation}
and if priors are identically distributed,
\begin{equation}
\label{eq_}
 =  H \mathbb{E} \big[   	w_{2}\psi(\mathbf{x}) w_{2}\psi(\mathbf{x}') 
   \big]
\end{equation}
\begin{equation}
\label{eq_deriv_last}
 =  \sigma_{w2}^2 \mathbb{E} \big[   	\psi(\mathbf{x}) \psi(\mathbf{x}') 
   \big]
\end{equation}
where $w_2 $ prior variance is scaled by width, $1/H$.  

Having derived expressions for mean and covariance, it remains to show that the distribution is Gaussian. Eq. \ref{eq_NN} is a sum of iid random variables, hence, under mild conditions, the CLT states that the distribution over functions is normally distributed as $H \to \infty$.

\subsection{ANALYTICAL BNN KERNELS}
\label{sec_analytical_bnn_kernels}
To derive analytical kernels for specific activations, $\psi$, and priors, $p(\mathbf{w}_1)$ \& $ p(b_1) $, eq. \ref{eq_deriv_last} must be evaluated.
\begin{equation}
\label{eq_kernel_int}
K(\mathbf{x}, \mathbf{x}') = \sigma_{w2}^2 \int \!\!\! \int  \psi( \mathbf{x}  )  \psi( \mathbf{x}' ) p(\mathbf{w}_1) p(b_1) d \mathbf{w}_1 d b_1
\end{equation}
The integral is generally not trivial, and several papers have focused on deriving analytical forms for popular activation functions, usually with normally distributed priors - ERF/probit (sigmoidal shape) and RBF \citep{Williams1996}, step function and ReLU \citep{Cho2009}, Leaky ReLU \citep{Tsuchida2018}. In section \ref{sec_periodic_BNNs} we add to this list by considering cosine activations. Similar results have been shown for convolutional BNNs \citep{Novak2018}.

Naturally eq. \ref{eq_kernel_int} can be computed numerically where analytical forms do not exist. Recurrent computation is necessary for deeper BNNs, which also converge to GPs \citep{Lee2018}.

\subsection{HYPERPARAMETER INTUITION}
\label{sec_vanil_BNNs}

Having shown a correspondence between GPs and BNNs, we now provide, in intuitive terms, the effect of key BNN hyperparameters on GP priors, which is useful when modelling with BNNs - care should then be taken to select hyperparameters that suit properties of the function being modelled.  We assume Gaussian priors on weights and biases, see \cite{Nalisnick2018} for an investigation of other prior distributions.

\begin{itemize}
\item \textbf{Activation function} - Swapping activations effectively swaps the parametric form of kernel. Basic BNNs in figure \ref{fig_priors} show example prior draws for single-layer BNNs with ReLU and ERF activations, as well as an RBF BNN.

\item \textbf{Prior variances} - These have different effects depending on the layer. Roughly speaking, variance of first layer weights and biases controls how wiggly the priors are (similar effect to length scale in the SE kernel). Final layer weight variance simply scales the output range of priors (similar effect to $\sigma^2$ in the SE kernel).


\item \textbf{Data noise variance} - A level of data noise variance (irreducible noise) must be specified to create a valid likelihood function when implementing BNNs. Normally distributed homoskedastic data noise is often assumed. Roughly speaking, data noise variance determines how perfectly the data should be fitted.


\end{itemize}

\section{KERNEL COMBINATIONS IN BNNs}
\label{sec_kernel_combos_BNN}
This section considers how to design BNN architectures such that, in the infinite width limit, they give rise to the equivalent GP kernel combinations.

The kernel combination operations we consider are; 
\begin{itemize}
\item \textbf{Addition}: $K(\mathbf{x},\mathbf{x}') = K_A(\mathbf{x},\mathbf{x}') + K_B(\mathbf{x},\mathbf{x}')$
\item \textbf{Multiplication}: $K(\mathbf{x},\mathbf{x}') = K_A(\mathbf{x},\mathbf{x}')  K_B(\mathbf{x},\mathbf{x}')$
\item \textbf{Polynomial}: e.g. $K(\mathbf{x},\mathbf{x}') = K_A(\mathbf{x},\mathbf{x}')^2 $
\item \textbf{Warping}: $K(\mathbf{x},\mathbf{x}') = K_A(u(\mathbf{x}),u(\mathbf{x}'))$ for a function, $u: R^d \to R^m$
\end{itemize}

We begin by considering architectures that combine the output of two BNNs. This turns out to be a valid way to add kernels, but not to multiply kernels. We then consider architectures that combine BNNs, point wise, at the final hidden layer. This is valid for multiplicative kernels, but produces a small artefact for additive kernels.

Having derived architectures mirroring additive and multiplicative kernels, section \ref{sec_extension} examines using these in more advanced ways.



\subsection{COMBINING BNNs AT OUTPUT}
\label{sec_combo_output}

A straightforward way to combine BNNs is to consider some operation combining their outputs.

\subsubsection{Additive}
\label{sec_output_add}
Consider two independent GPs denoted $f_A(\mathbf{x})$ \& $f_B(\mathbf{x})$, summed,
\begin{equation}
\label{eq_BNN_add_output_GP}
f_{add}(\mathbf{x}) = f_A(\mathbf{x}) + f_B(\mathbf{x}).
\end{equation}
In general, it is known that $f_{add}(\mathbf{x})$ will also be a GP with kernel, $K_{add}(\mathbf{x},\mathbf{x}') = K_A(\mathbf{x},\mathbf{x}') + K_B(\mathbf{x},\mathbf{x}')$, \citep{Saul2016}. 

For two single-layer BNNs, this is recovered by a BNN of architecture,
\begin{equation}
\label{eq_BNN_add_output_BNN}
= \sum_{i=1}^H w_{A2i}   \psi_{Ai}(\mathbf{x} ) + \sum_{j=1}^H w_{B2j}   \psi_{Bj}(\mathbf{x} ).
\end{equation}
Since this converges to the sum of two independent GPs, regardless of depth (section \ref{sec_one_act}), the general GP result applies, and suffices to show that independent BNNs (of infinite width) summed at outputs reproduce a GP with additive kernel.

\subsubsection{Multiplicative}

Two GPs multiplied together, 
\begin{equation}
\label{eq_mult_output_GP}
f_{mult}(\mathbf{x}) = f_A(\mathbf{x}) f_B(\mathbf{x}),
\end{equation}
do \textit{not} generally produce a GP \citep{Rasmussen2018} [4.2.4], even though there does exist a GP with kernel, $K_{mult}(\mathbf{x},\mathbf{x}') = K_A(\mathbf{x},\mathbf{x}') K_B(\mathbf{x},\mathbf{x}')$. (Analogously, the product of two normally distributed random variables is not normally distributed.)

This means that independent BNNs multipled at outputs (shown for the single layer case),
\begin{equation}
\label{eq_mult_output_BNN}
= \sum_{i=1}^H w_{A2i}   \psi_{Ai}(\mathbf{x} ) \sum_{j=1}^H w_{B2j}   \psi_{Bj}(\mathbf{x} )
\end{equation}
do \textit{not} produce a GP with multiplied kernel.

\subsection{COMBINING BNNs AT HIDDEN LAYERS}

Consider now combining BNNs by point wise operations at their hidden layers. 

\subsubsection{Additive}

Taking two single-layer BNNs, the additive case is,
\begin{equation}
\label{eq_BNN_add_hidden}
f(\mathbf{x}) = \sum_{i=1}^H w_{2i}   \big( \psi_{Ai}(\mathbf{x} ) + \psi_{Bi}(\mathbf{x} ) \big),
\end{equation}
where $\psi_{A}$ and $\psi_{B}$ are hidden units for each sub-BNN. As in section \ref{sec_combo_output}, neither hyperparameters nor activation function need be shared, e.g. one could take a RBF and ReLU BNN, $\psi_{A}(\mathbf{x}) = \exp ( - \lvert \lvert \mathbf{x} - \mathbf{w}_{A1}^T\rvert \rvert_2^2 /\sigma_g^2)$, and, $\psi_{B}(\mathbf{x}) = \max (\mathbf{w}_{B1} \mathbf{x}+ b_{B1},0)$.
We now derive the equivalent GP for such an architecture. 

Analysis precisely as in section \ref{sec_one_act} can be followed up to eq. \ref{eq_deriv_last}, leaving,
\begin{equation}
\label{eq_}
\begin{split}
K_{add}(\mathbf{x},\mathbf{x}') = &  \\  \sigma_{w2}^2  \mathbb{E} \big[   \big( &  \psi_{A}(\mathbf{x} ) + \psi_{B}(\mathbf{x} ) \big) \big( \psi_{A}(\mathbf{x}' ) + \psi_{B}(\mathbf{x}' ) \big) 
   \big]
\end{split}
\end{equation}
\begin{equation}
\begin{split}
\label{eq_}
=  \sigma_{w2}^2 \mathbb{E} \big[   \psi_{A}(\mathbf{x} ) \psi_{A}(\mathbf{x}' ) +
																	\psi_{A}(\mathbf{x} ) \psi_{B}(\mathbf{x}' ) + \\
																	\psi_{B}(\mathbf{x} ) \psi_{A}(\mathbf{x}' ) +
																	\psi_{B}(\mathbf{x} ) \psi_{B}(\mathbf{x}' )   \big]
\end{split}
\end{equation}
by linearity of expectation, and noting $\psi_{A}$ and $\psi_{B}$ are independent,
\begin{equation}
\begin{split}
\label{eq_}
= & \sigma_{w2}^2 \mathbb{E} \big[   \psi_{A}(\mathbf{x} ) \psi_{A}(\mathbf{x}' )] +
     \sigma_{w2}^2 \mathbb{E} \big[	\psi_{B}(\mathbf{x} ) \psi_{B}(\mathbf{x}' )     \big] +  \\
    & \sigma_{w2}^2 \mathbb{E} \big[	\psi_{A}(\mathbf{x} ) \big]  \mathbb{E} \big[ \psi_{B}(\mathbf{x}' )     \big] + 
     \sigma_{w2}^2 \mathbb{E} \big[	\psi_{A}(\mathbf{x}' ) ]  \mathbb{E} \big[ \psi_{B}(\mathbf{x} )     \big]
\end{split}
\end{equation}
\begin{equation}
\begin{split}
\label{eq_}
=  K_A(\mathbf{x},\mathbf{x}') + K_B(\mathbf{x},\mathbf{x}') & + \\ 
	\sigma_{w2}^2 \mathbb{E} \big[	\psi_{A}(\mathbf{x} ) \big]  \mathbb{E} \big[ \psi_{B}(\mathbf{x}' )   &  \big] + 
     \sigma_{w2}^2 \mathbb{E} \big[	\psi_{A}(\mathbf{x}' ) ]  \mathbb{E} \big[ \psi_{B}(\mathbf{x} )     \big].
\end{split}
\end{equation}
This is the additive kernel plus two extra terms. The impact of these extra terms depends on the activation function, and could be compensated for. If either $\psi$ is an odd function, $ \mathbb{E} \big[	\psi_{odd}(\mathbf{\cdot}) ] = 0$, the additive kernel is exactly recovered. Alternatively, if both $\psi$'s are sigmoids, $ \mathbb{E} \big[	\psi_{sig}(\mathbf{\cdot}) ] = 0.5$, which results in the kernel, $ K_A(\mathbf{x},\mathbf{x}') + K_B(\mathbf{x},\mathbf{x}') + c$, for some constant $c$. If $\psi$ is a ReLU, $ \mathbb{E} \big[	\psi_{ReLU}(\mathbf{\cdot}) ] $ is  input dependent, making compensation trickier (though still possible).

In general, summing point wise after hidden nodes is not a valid way to reproduce an additive GP kernel, although effects of the artefact terms could be compensated for.

\subsubsection{Multiplicative}

Following the same procedure for multiplication after hidden nodes,
\begin{equation}
\label{eq_BNN_mult_hidden}
f(\mathbf{x}) = \sum_{i=1}^H w_{2i}   \big( \psi_{Ai}(\mathbf{x} )  \psi_{Bi}(\mathbf{x} ) \big)
\end{equation}
\begin{equation}
\label{eq_}
\begin{split}
K_{mult}(\mathbf{x},\mathbf{x}') = &  \\  \sigma_{w2}^2  \mathbb{E} & \big[   \big( \psi_{A}(\mathbf{x} )  \psi_{B}(\mathbf{x} ) \big)  \big( \psi_{A}(\mathbf{x}' )  \psi_{B}(\mathbf{x}' ) \big) 
   \big]
\end{split}
\end{equation}
BNN independence allows the rearrangement,
\begin{equation}
\label{eq_BNN_mult_derive}
=  \sigma_{w2}^2  \mathbb{E} \big[   \psi_{A}(\mathbf{x} )  \psi_{A}(\mathbf{x}' )   \big]
   					   \mathbb{E} \big[  \psi_{B}(\mathbf{x} )  \psi_{B}(\mathbf{x}' )  \big]
\end{equation}
\begin{equation}
\label{eq_}
 =  K_A(\mathbf{x},\mathbf{x}')  K_B(\mathbf{x},\mathbf{x}')
\end{equation}
and hence multiplying point wise after hidden nodes is a valid way to reproduce a multiplicative GP kernel.

\subsection{EXTENSIONS}
\label{sec_extension}
Whilst the previous results were explicitly shown for two single-layer BNNs, it is straightforward to extend them to a variety of situations. Following, we provide examples of useful constructions.

\textbf{Additive and Multiplicative}

Kernel:
\begin{equation}
\label{eq_}
\begin{split}
K_{}(\mathbf{x},\mathbf{x}')  =  K_A(\mathbf{x},\mathbf{x}')  + K_B(\mathbf{x},\mathbf{x}')  K_C(\mathbf{x},\mathbf{x}')  K_D(\mathbf{x},\mathbf{x}') 
\end{split}
\end{equation}
\vskip -0.1in
BNN architecture:
\begin{equation}
\label{eq_}
f(\mathbf{x}) = 
\sum_{i=1}^H w_{2i}   \psi_{Ai}(\mathbf{x} ) +
\sum_{j=1}^H w_{2j} \psi_{Bj}(\mathbf{x})  \psi_{Cj}(\mathbf{x}) \psi_{Dj}(\mathbf{x} )
\end{equation}

\textbf{Polynomials}

Kernel:
\begin{equation}
\label{eq_}
K_{}(\mathbf{x},\mathbf{x}')  =  K_A(\mathbf{x},\mathbf{x}')^2
\end{equation}
\vskip -0.1in
BNN architecture:
\begin{equation}
\label{eq_}
f(\mathbf{x}) = 
\sum_{i=1}^H w_{2i}   \psi_{A1i}(\mathbf{x} )  \psi_{A2i}(\mathbf{x})
\end{equation}
Where $\psi_{A1}$ and $\psi_{A2}$ are separate nodes sharing common hyperparameters. 

\textbf{Warping}


Kernel:
\begin{equation}
K_{}(\mathbf{x},\mathbf{x}')  = K(  u(\mathbf{x}), u(\mathbf{x}')  )
\end{equation}
\vskip -0.1in
BNN architecture:
\begin{equation}
\label{eq_}
f(\mathbf{x}) = 
\sum_{i=1}^H w_{2i}  \psi_i(u(\mathbf{x}))
\end{equation}

\begin{figure}[t!]
\begin{center}
	 \vskip -0.1in
	 
    \begin{minipage}{\widthPriors}
    	\centering
        \begin{overpic}[width=\linewidth]{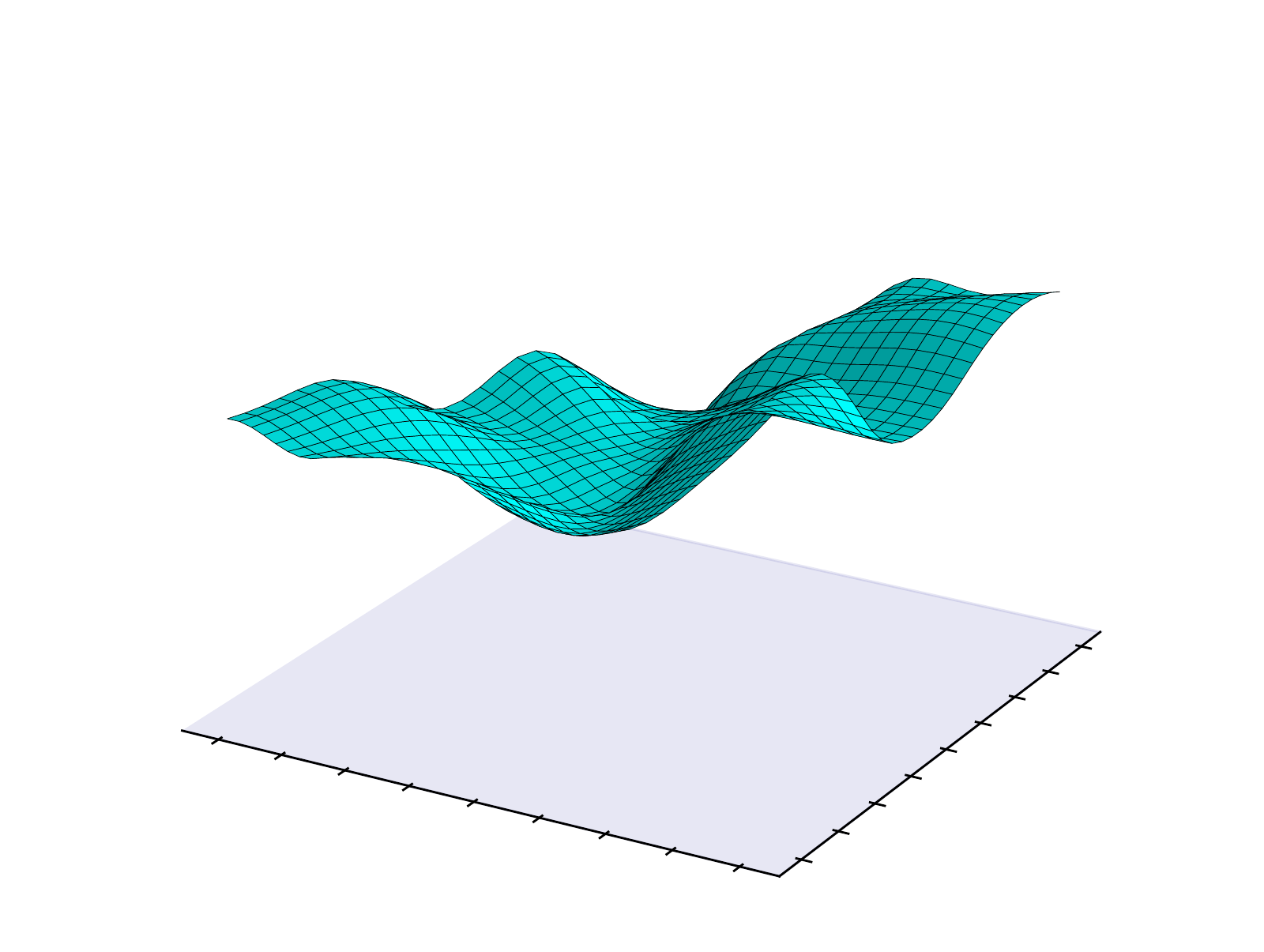}
         \put(50,9){\makebox(0,0){ \small{Basic BNN, ERF$[0:1]$}}}
        \end{overpic}
    \end{minipage} \hskip -0.05in
    \begin{minipage}{\widthPriors}
    	\centering
        \begin{overpic}[width=\linewidth]{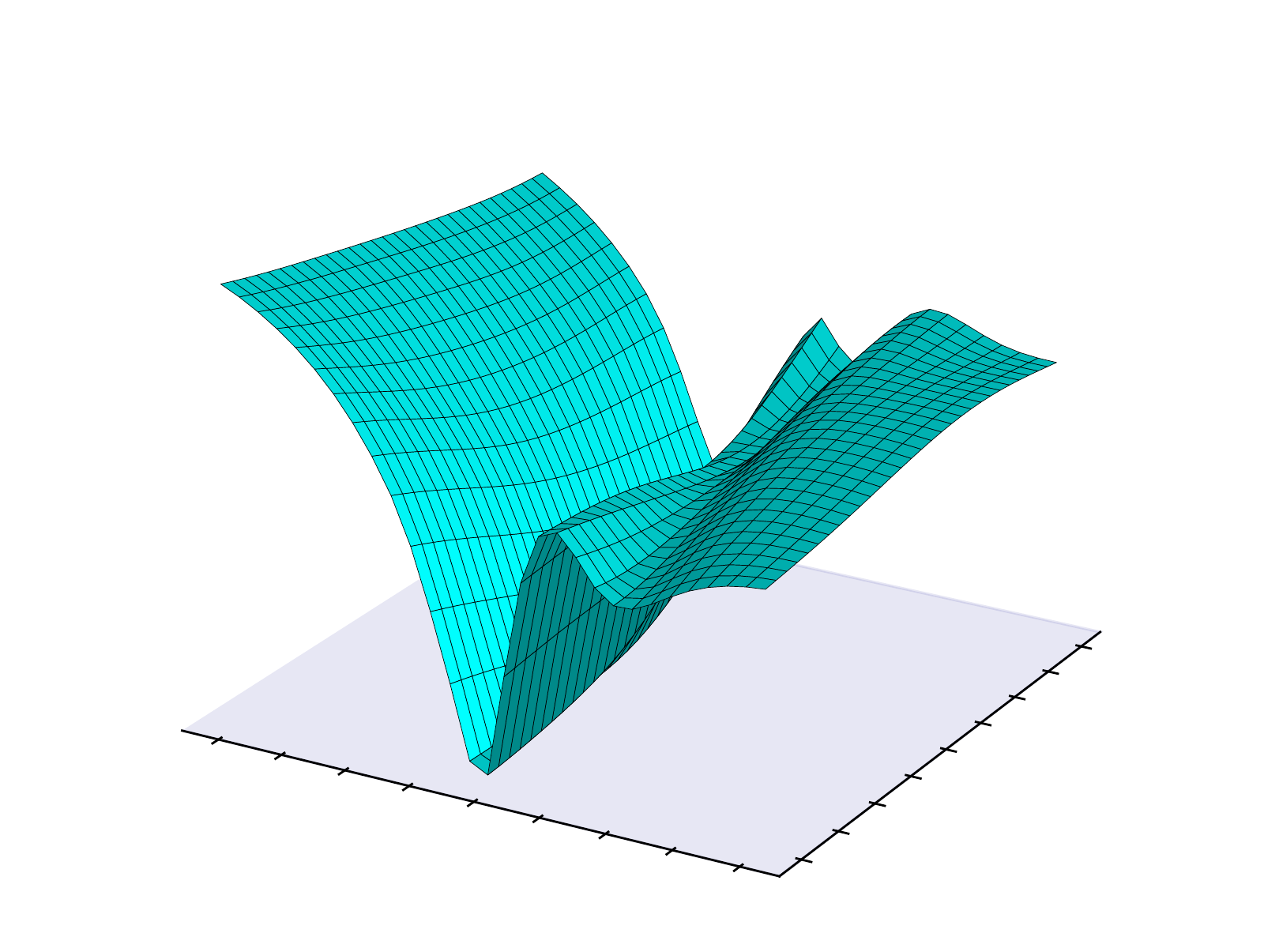} 
        \put(50,9){\makebox(0,0){ \small{ERF$[0]$ $+$ ERF$[1]$}}}
        \end{overpic}
    \end{minipage}
   
     \vskip -0.1in
     
    \begin{minipage}{\widthPriors}
    	\centering

        \begin{overpic}[width=\linewidth ]{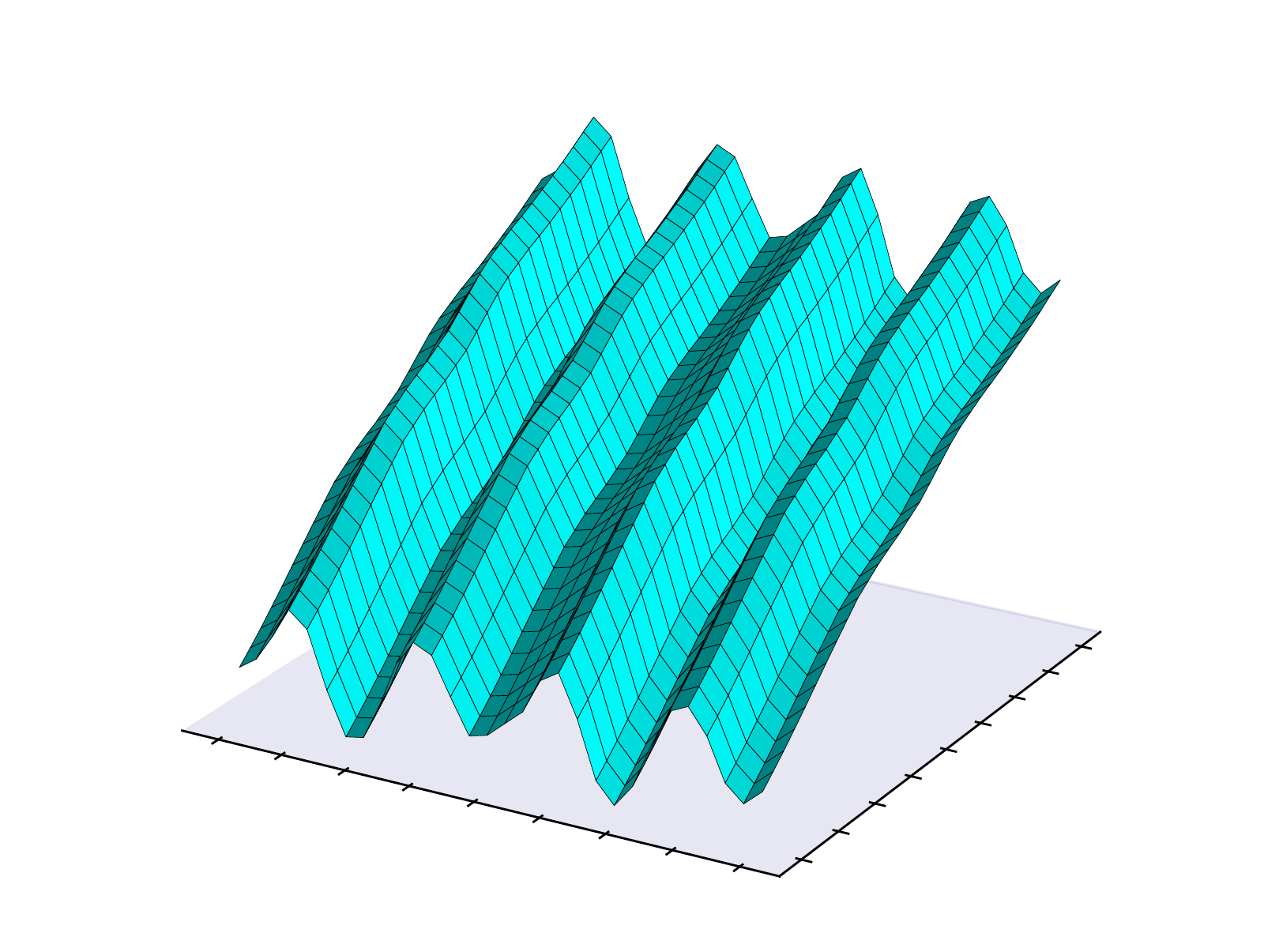}
         \put(50,9){\makebox(0,0){ \small{Periodic$[0] +$ ReLU$[1]$}}}
        \end{overpic}
    \end{minipage} \hskip -0.05in
    \begin{minipage}{\widthPriors}
    	\centering
        \begin{overpic}[width=\linewidth]{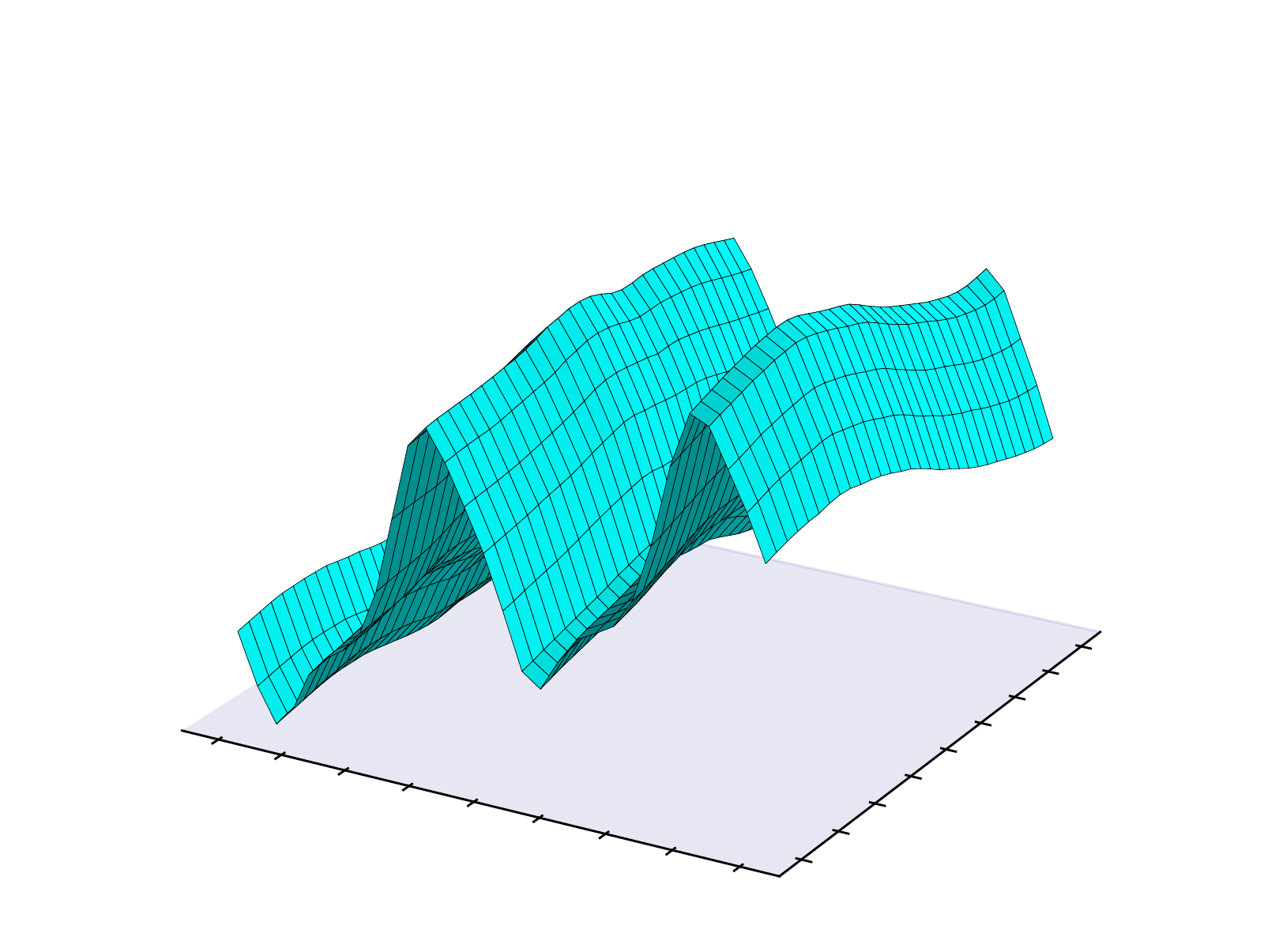}
        \put(50,9){\makebox(0,0){ \small{Periodic$[0] \times$ ReLU$[0:1]$}}}
        \end{overpic}
    \end{minipage}
\vskip -0.05in
\caption{Prior draws for a $2$-$D$ input. Square brackets designate which dimension(s) each kernel is applied to.}
\label{fig_priors_3d}
\end{center}
\vskip -0.1in
\end{figure}

\textbf{Separation of Inputs}

It can be useful to consider multiple kernels taking subsets of inputs, combined through either addition or multiplication \citep{Duvenaud2014} [2.3, 2.4], as visualised in figure \ref{fig_priors_3d}.

Kernel:
\begin{equation}
\label{eq_}
K( \mathbf{x}, \mathbf{x}' ) = K_A( x_1, x_1' ) +  K_B( x_2, x_2' ) 
\end{equation}
\vskip -0.1in
BNN architecture:
\begin{equation}
\label{eq_}
f(\mathbf{x}) = 
\sum_{i=1}^H w_{2i}  \psi_{Ai}(x_1) + \sum_{j=1}^H w_{2j}  \psi_{Bj}(x_2)
\end{equation}

Kernel:
\begin{equation}
\label{eq_}
K( \mathbf{x}, \mathbf{x}' ) = K_A( x_1, x_1' )   K_B( x_2, x_2' ) .
\end{equation}
\vskip -0.1in
BNN architecture: 
\begin{equation}
\label{eq_}
f(\mathbf{x}) = 
\sum_{i=1}^H w_{2i}  \psi_{Ai}(x_1)  \psi_{Bi}(x_2).
\end{equation}

%

\textbf{Deeper BNNs}

Out of convenience, constructions have been shown for single-layer BNNs. These could be replaced by deep BNNs, which equally correspond to GP kernels (section \ref{sec_analytical_bnn_kernels}).

\section{PERIODIC BNN KERNELS}
\label{sec_periodic_BNNs}
This section considers how BNNs can be designed to model periodic functions. To our knowledge this analysis is entirely novel. We define a periodic function as, $f(x) = f(x+p)$, for some scalar period, $p \in \mathbb{R}_+$.

We find that cosine activations do not produce a periodic kernel, but applying warping to inputs followed by standard activations functions, does.

\subsection{COSINE ACTIVATIONS}

Consider a single-layer BNN with cosine activation functions; intuition might suggest this leads to a periodic kernel. (Note such activations have been explored in other contexts \citep{Parascandolo2017, Ramachandran2017}.)
\begin{equation}
\label{eq_}
f(\mathbf{x}) = \sum_{i=1}^H w_{2i}    \cos( \mathbf{w}_{1i} \mathbf{x}+ b_{1i} )
\end{equation}
Following the usual GP kernel derivation in section \ref{sec_one_act},
\begin{equation}
\label{eq_}
\begin{split}
&K_{cos}(\mathbf{x}, \mathbf{x}') =  \\
& \sigma_{w2}^2 \int \!\!\! \int \cos( \mathbf{w}_{1} \mathbf{x}+ b_{1} ) \cos( \mathbf{w}_{1} \mathbf{x}'+ b_{1} )  p(\mathbf{w}_1) p(b_1) d \mathbf{w}_1 d b_1
\end{split}
\end{equation}
Assuming priors, $p(\mathbf{w}_1) \sim \mathcal{N}(0,\sigma^2_{w1}I)$, and, $p(b_1) \sim \mathcal{N}(0,\sigma^2_{b1})$, we find,\footnote{Rewrite $\cos(A) \cos(B) = \frac{1}{2} [\cos(A-B) + \cos(A+B)]$, then use, $ \mathbb{E} [ \cos(\mathbf{xw})  ]=  \exp(-\frac{1}{2} \mathbf{x}^T \Sigma \mathbf{x} ) $, if  $\mathbf{w} \sim \mathcal{N}(0,\Sigma)$.}

\begin{equation}
\label{eq_}
= \frac{\sigma^2_{w2}}{2}\bigg(   \exp  \Big(  {- \frac{\lvert \lvert \mathbf{x} - \mathbf{x}'\rvert \rvert^2_2 }  {2 / \sigma^2_{w1} }} \Big)
  + \exp  \Big( {- \frac{\lvert \lvert \mathbf{x} + \mathbf{x}'\rvert \rvert^2_2 }  {2 / \sigma^2_{w1} }  +  2 \sigma^2_{b1}}  \Big) \bigg).
\end{equation}
Slightly counter-intuitively, the kernel is not periodic. Rather it is the sum of the SE kernel (eq. \ref{eq_SE_GP}), and another term.

We further considered using Laplace and uniform distributions for priors, which did result in kernels containing trigonometric functions, but the forms were untidy and not apparently useful.

Note that our analysis is from the perspective of equivalent GP kernels. It is possible to consider narrow BNNs with cosine activations that \textit{would} produce periodic predictive distributions. If initialised suitably, these may be of some use.


\newcommand\dist{72} 
\newcommand\distw{3}
\newcommand\widthTimeSeries{0.24}
\begin{figure*}[t!]
\begin{center}

    \begin{minipage}{\widthTimeSeries\textwidth}
        \centering
        \begin{overpic}[width=1.\textwidth]{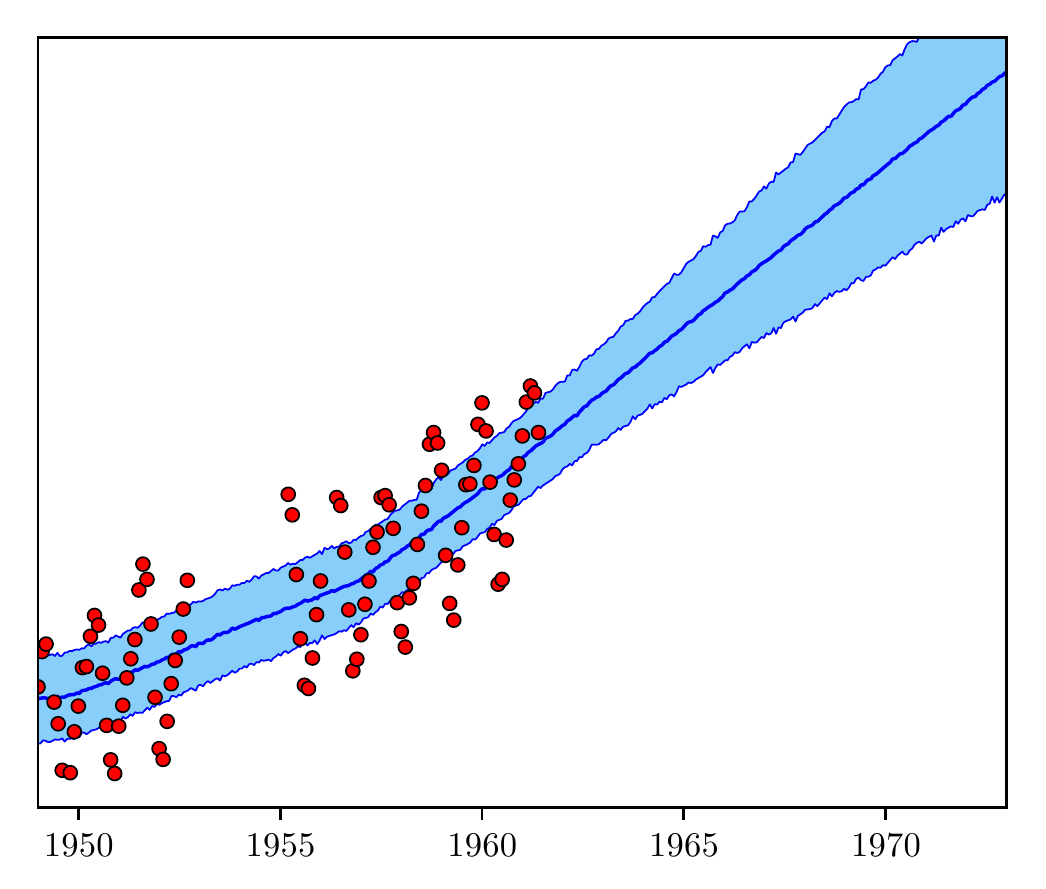}
         \put(-8,16){\rotatebox{90}{Mauna Loa CO$_2$}}
         \put(\distw,\dist){ \small{ReLU BNN}}
        \end{overpic}
    \end{minipage}
    \begin{minipage}{\widthTimeSeries\textwidth}
        \centering
        \begin{overpic}[width=1.\textwidth]{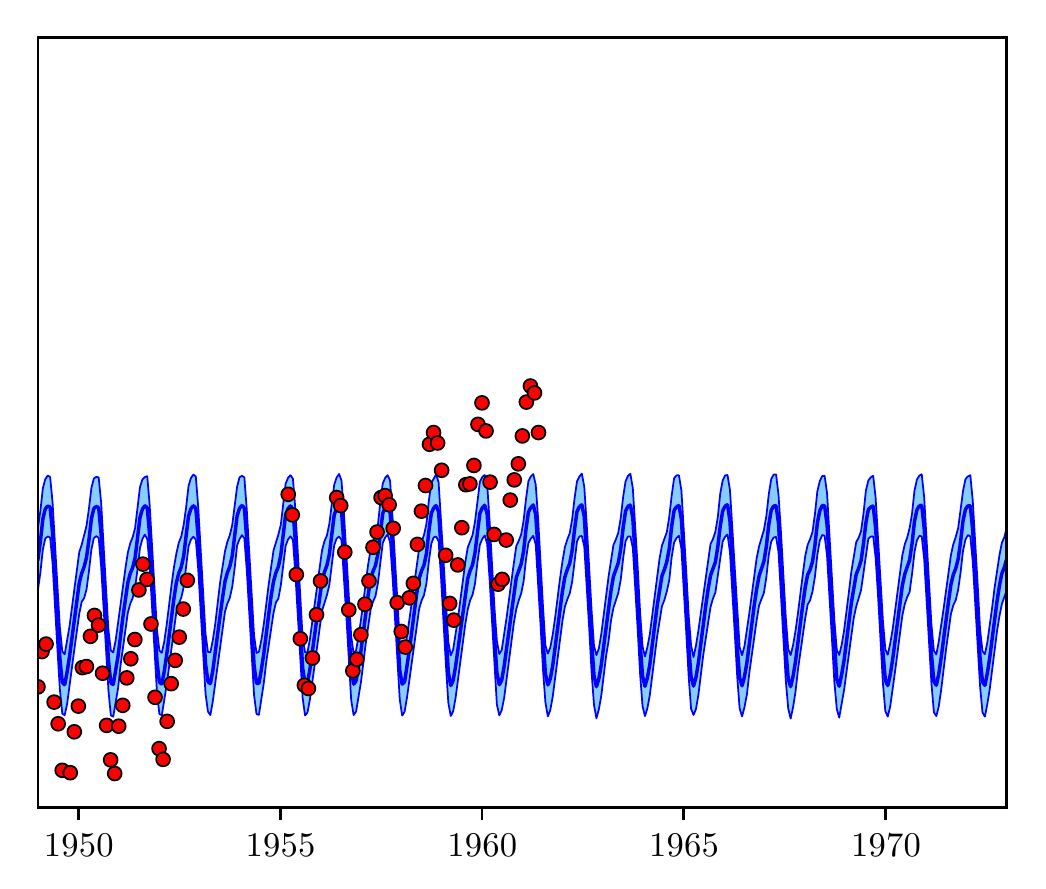}
         \put(\distw,\dist){ \small{Periodic BNN}}
        \end{overpic}
    \end{minipage}
    \begin{minipage}{\widthTimeSeries\textwidth}
        \centering
        \begin{overpic}[width=1.\textwidth]{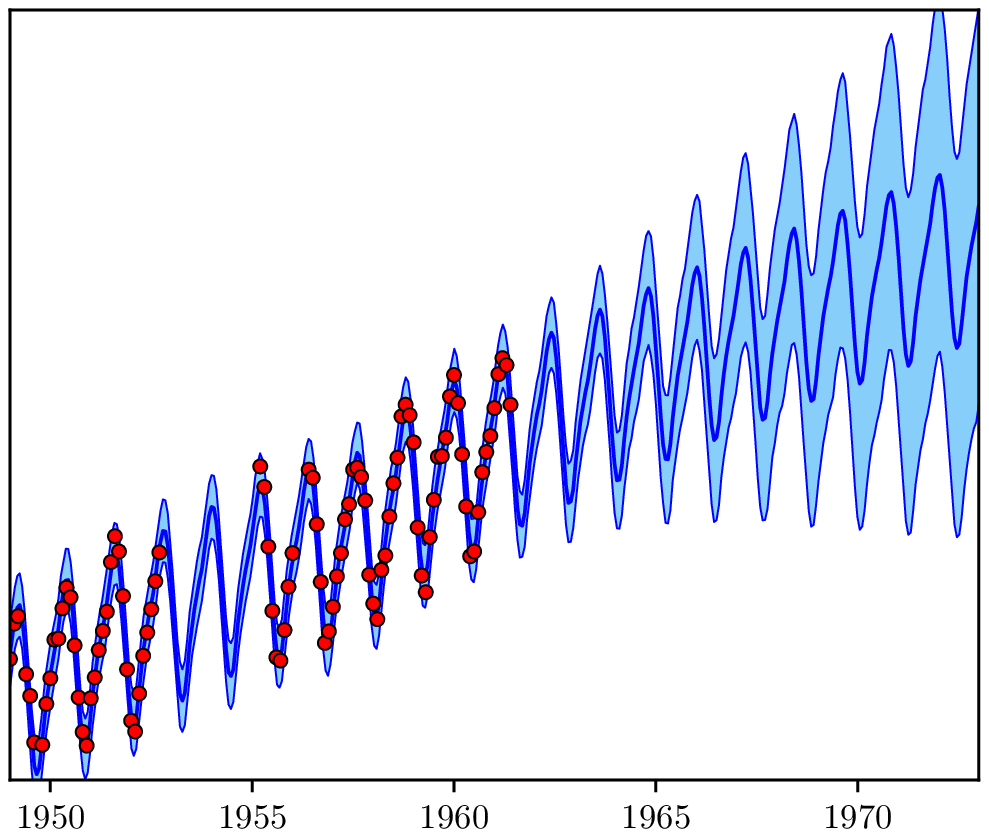}
         \put(\distw,\dist){ \small{ReLU $+$ Periodic BNN}}
        \end{overpic}
    \end{minipage}
    \begin{minipage}{\widthTimeSeries\textwidth}
        \centering
        \begin{overpic}[width=1.\textwidth]{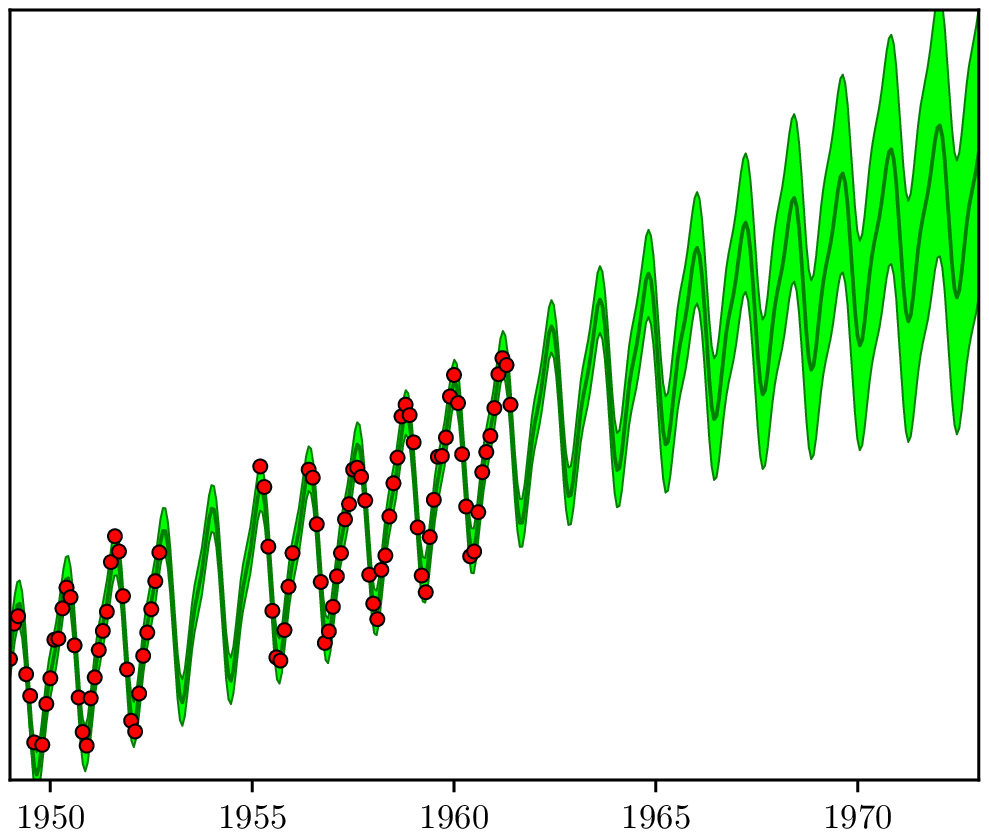}
         \put(\distw,\dist){ \small{ReLU $+$ Periodic GP}}
        \end{overpic}
    \end{minipage}
    
    \vskip -0.02in

    \begin{minipage}{\widthTimeSeries\textwidth}
        \centering
        \begin{overpic}[width=1.\textwidth]{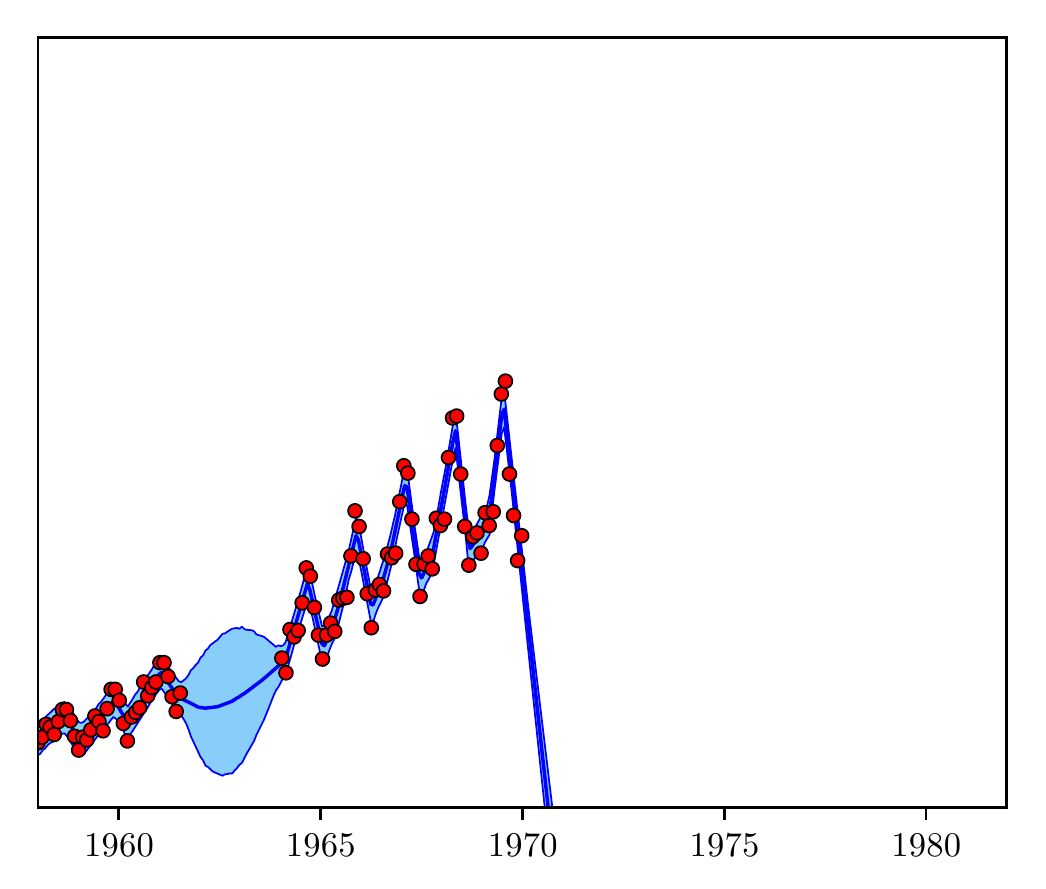}
         \put(-8,10){\rotatebox{90}{Airline Passengers}}
         \put(\distw,\dist){ \small{ReLU BNN}}
        \end{overpic}
    \end{minipage}
    \begin{minipage}{\widthTimeSeries\textwidth}
        \centering
        \begin{overpic}[width=1.\textwidth]{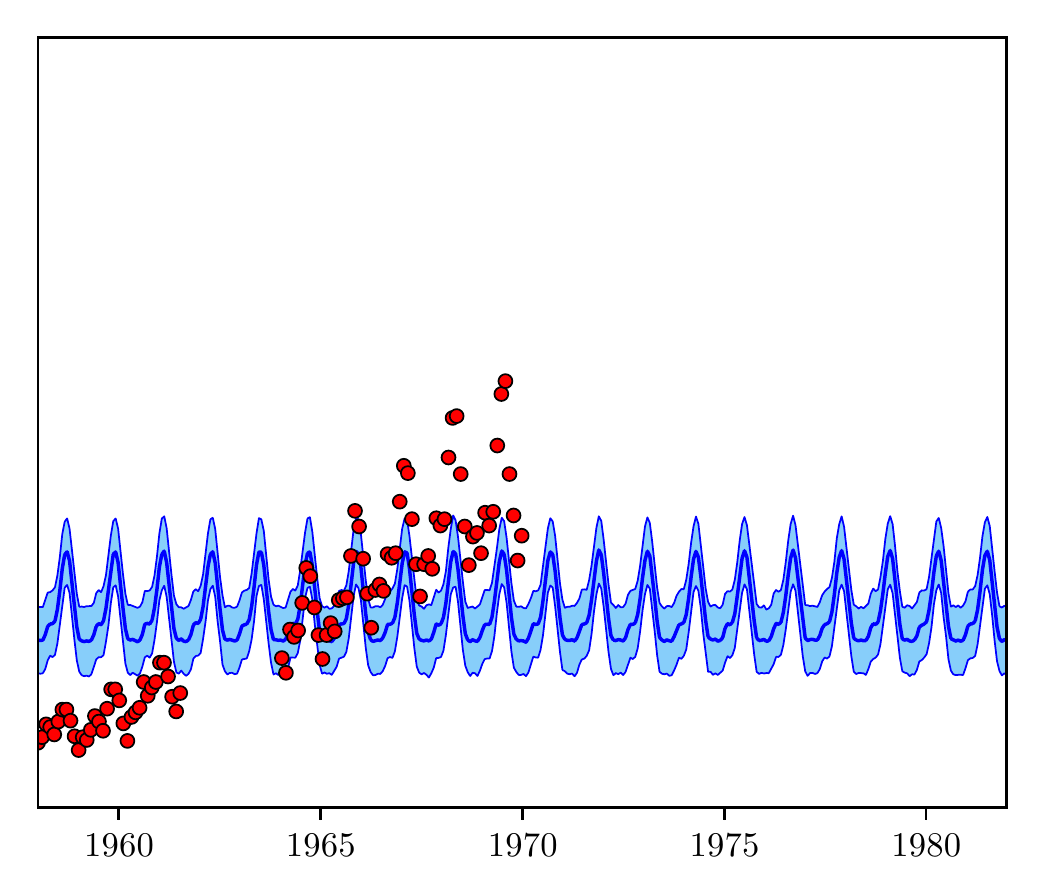}
         \put(\distw,\dist){ \small{Periodic BNN}}
        \end{overpic}
    \end{minipage}
    \begin{minipage}{\widthTimeSeries\textwidth}
        \centering
        \begin{overpic}[width=1.\textwidth]{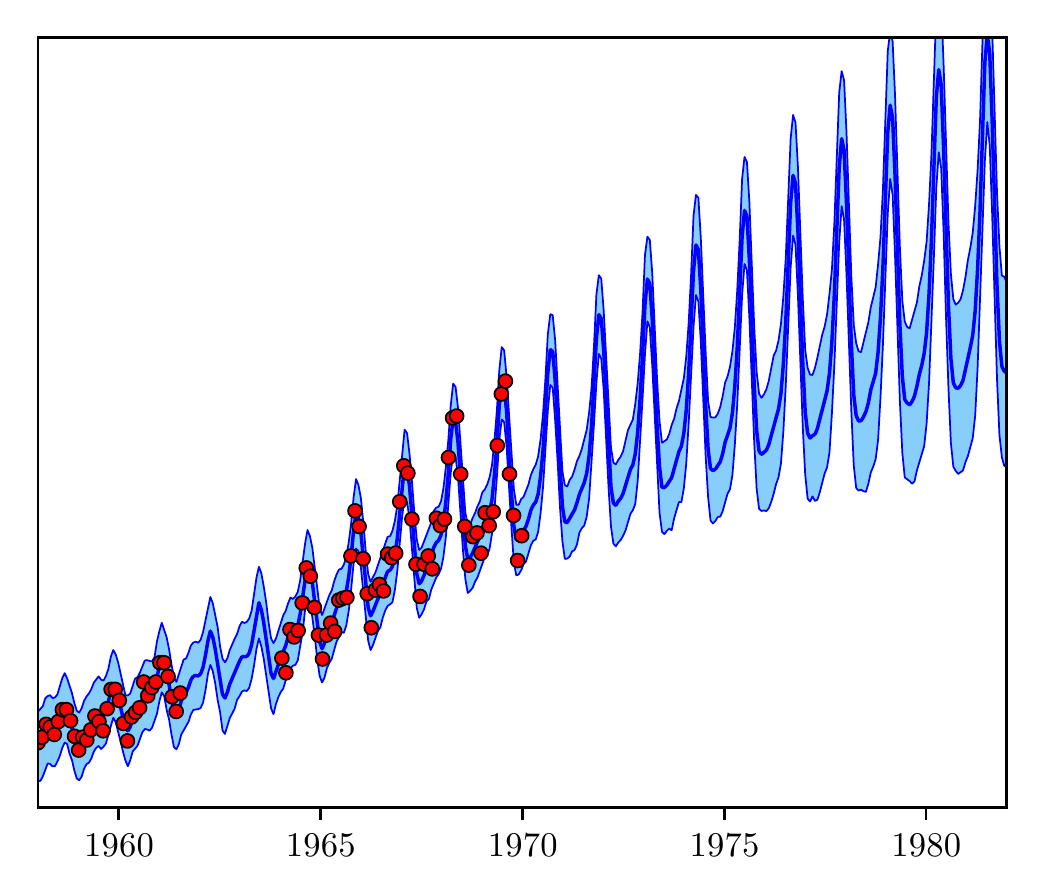}
         \put(\distw,\dist){ \small{ReLU $\times$ Periodic BNN}}
        \end{overpic}
    \end{minipage}
    \begin{minipage}{\widthTimeSeries\textwidth}
        \centering
        \begin{overpic}[width=1.\textwidth]{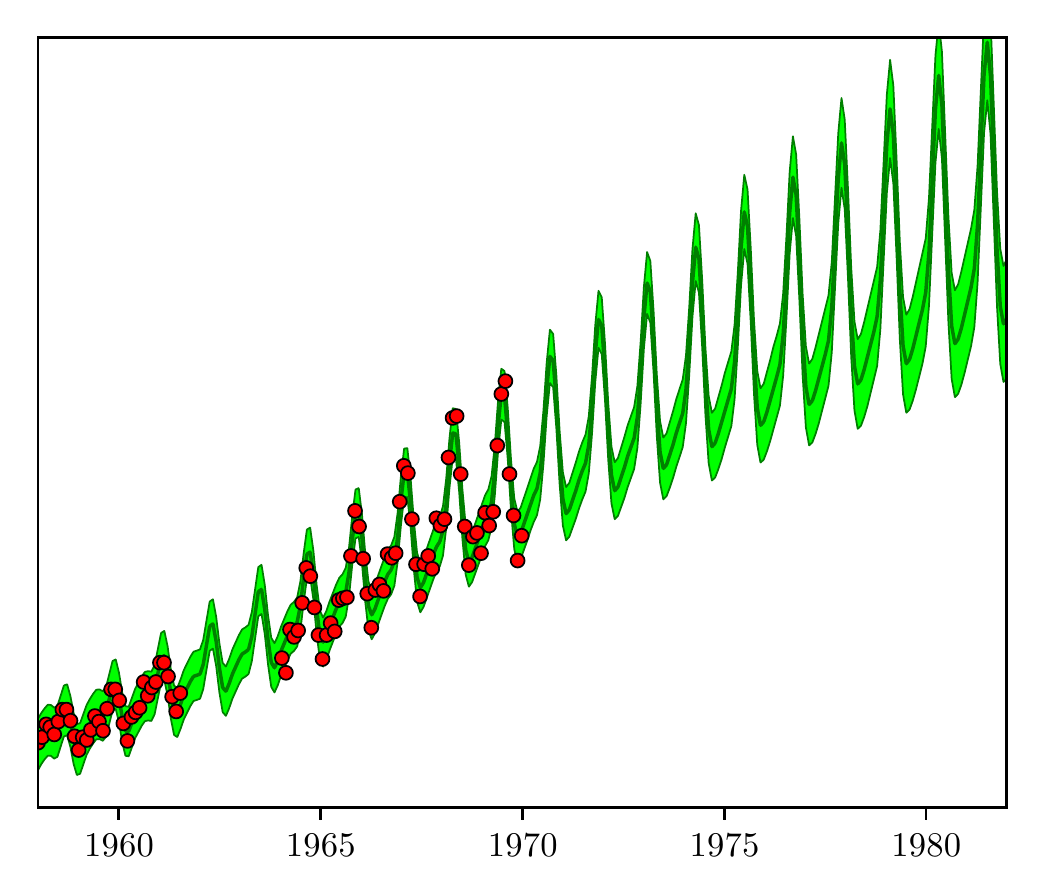}
         \put(\distw,\dist){ \small{ReLU $\times$ Periodic GP}}
        \end{overpic}
    \end{minipage}
\caption{Two time series with seasonal fluctuations and long term trends. ReLU and periodic BNNs are separately unable to capture these patterns (first and second from left). However, they succeed when combined in BNN architectures as proposed in this paper (third from left), closely approximating the predictive distributions of exact GP inference with the equivalent kernel combinations (right).}
\label{fig_timeseries}
\end{center}
\vskip -0.2in
\end{figure*}

\subsection{INPUT WARPING}

Whilst modifying the activation function failed to produce periodic kernels, applying a warping to inputs was more successful.

The most common periodic kernel used in GP modelling is the ESS kernel \citep{Duvenaud2014} [p. 25], as shown in eq. \ref{eq_GPperiodic}. Having established its value in periodic modelling, we wanted to reproduce this as closely as possible with a BNN. Surprisingly, an exact recovery is possible as follows.

Apply a warping to a 1-D input, $x \to (\cos( \frac{2 \pi x}{p} ),  \sin( \frac{2 \pi x}{p} ))$, followed by a single-layer RBF BNN taking this 2-D warping as input. 

In general, an infinitely wide single-layer RBF BNN produces the following GP kernel \citep{Williams1996},
\begin{equation}
\label{eq_rbf_bnn_kernel}
\begin{split}
K_{RBF_{BNN}}&(\mathbf{x},\mathbf{x}') = \bigg( \frac{\sigma_e}{\sigma_u}\bigg)^d 
\exp \bigg( - \frac{\mathbf{x}^T\mathbf{x}}{2\sigma^2_m} \bigg) \\
&\exp \bigg( - \frac{ \lvert \lvert \mathbf{x} - \mathbf{x}' \rvert  \rvert_2 ^2 }{2\sigma^2_s} \bigg)
\exp \bigg( - \frac{\mathbf{x}'^T\mathbf{x}'}{2\sigma^2_m} \bigg)
\end{split}
\end{equation}
where, $1/\sigma^2_e = 2/\sigma^2_g + 1/\sigma^2_u$, $\sigma^2_s = 2\sigma^2_g + \sigma^4_g / \sigma^2_u$, and $\sigma^2_m = 2\sigma^2_u + \sigma^2_g$. If the discussed warping is first applied, for the 1-D case this becomes,
\begin{equation}
\begin{split}
 K_{RBF Per_{BNN}}&(\mathbf{x},\mathbf{x}') = \\
 \bigg( \frac{\sigma_e}{\sigma_u}\bigg)^2 
& \exp \bigg( - \frac{ \cos^2( \frac{2 \pi x}{p} ) + \sin^2( \frac{2 \pi x}{p} )  }{2\sigma^2_m} \bigg)  \\
& \exp \bigg( - \frac{  \big( \cos( \frac{2 \pi x}{p} )  - \cos( \frac{2 \pi x'}{p} ) \big)^2}{2\sigma^2_s} \\
 &   \;\;\;\;\;\;\;\; +  \frac{  \big( \sin( \frac{2 \pi x}{p} )  - \sin( \frac{2 \pi x'}{p} ) \big)^2     }{2\sigma^2_s} \bigg)  \\
& \exp \bigg( - \frac{ \cos^2( \frac{2 \pi x'}{p} ) + \sin^2( \frac{2 \pi x'}{p} )  }{2\sigma^2_m} \bigg).
\end{split}
\end{equation}
Noting, $\big(\cos( \frac{2 \pi x}{p} )  - \cos( \frac{2 \pi x'}{p} ) \big)^2 + \big( \sin( \frac{2 \pi x}{p} )  - \sin( \frac{2 \pi x'}{p} ) \big)^2 = 4 \sin^2(\frac{ \pi }{p} (x-x') )$, and also, $\cos^2( \cdot ) + \sin^2( \cdot ) = 1$, this reduces to,
\begin{equation}
\label{eq_BNN_per_RBF}
= \bigg( \frac{\sigma_e}{\sigma_u}\bigg)^2
\exp \bigg( - \frac{1}{\sigma^2_m} \bigg)
\exp \bigg( -  \frac{2 \sin^2(\frac{ \pi }{p} (x-x') ) }{\sigma^2_s}   \bigg)
\end{equation}
which is of the same form as the periodic ESS kernel. Indeed there is a connection to the derivation of the ESS kernel, which used the same warping followed by the SE kernel \citep{mackay1998}. 


It is equally plausible to apply the same warping followed by BNNs of other architectures. For example, the single-layer ReLU case results in,
\begin{equation}
\label{eq_BNN_relu_per}
K_{ReLU Per}(\mathbf{x},\mathbf{x}') =  \frac{\sigma^2_{w_2}}{\pi} (\sin \omega + (\pi - \omega)\cos \omega)
\end{equation}
where,
\begin{equation}
\omega = \cos^{-1} \bigg( 
\frac{\sigma^2_{b_1} + \sigma^2_{w_1}  \cos(\frac{2 \pi}{p} (x - x')) }
{ \sigma^2_{b_1} + \sigma^2_{w_1}  } 
\bigg).
\end{equation}
This is equally suited to periodic modelling, and perhaps more convenient in BNNs given the prevalence of ReLUs.

\section{ILLUSTRATIVE EXPERIMENTS}
\label{sec_experiments}

This section provides examples of where, all things being equal, BNNs designed to incorporate suitable prior knowledge can deliver a performance boost over basic BNNs. These gains should be independent of learning algorithm or inference method, but are necessarily task specific. Hence, experiments are framed as illustrative rather than exhaustive.

We showcase situations benefiting simultaneously from both of the ideas introduced in this paper - combinations of BNNs \textit{and} periodic function modelling, although either can also be used separately.

All experiments used BNN widths of 50 hidden nodes. Their success supports our claim that, despite the theory presented in this paper being exact only for infinite-width BNNs, it provides sound principles for building expressive BNN models of finite width.

\subsection{SUPERVISED LEARNING: TIME SERIES}

Time series data often have seasonal fluctuations combined with longer term trends. These experiments show that where basic BNNs struggle to capture such patterns, simple combinations of these basic BNNs produce appropriate priors.


We considered two prediction tasks: CO$_2$ levels recorded at a Hawaiian volcano (Mauna), and numbers of airline passengers flying internationally (Airline). For both datasets we used ten years of monthly recordings, then deleted data between years 3-5 to create a gap in the series. Below, we qualitatively assess the predictive distribution in both the interpolation region (3-5 years) and an extrapolation region (10-20 years).

In Mauna, seasonal variations appear to be of constant amplitude, suggesting an additive relationship between trend and period, whilst Airline shows increasing amplitudes, suggesting a multiplicative relationship.

Figure \ref{fig_timeseries} shows the two datasets and the predictive distributions produced by four types of model (shading gives $\pm$3 standard deviations). Inference was performed with HMC for BNNs \citep{Neal1997}, and analytically for GP.
\begin{enumerate}
\item \textbf{ReLU BNN} - single-layer BNN with ReLU activations. There are two possibilities with this model - a long length scale, as shown for Mauna, which captures the long term trend but does not fit the seasonal variations. Alternatively, a short length scale allows better fitting of the training data, but at the expense of extrapolations - in Airline this produces a nonsensical 10-20 year forecast.

\item \textbf{Periodic BNN} - single-layer BNN with cos/sin warping applied, followed by RBF activations. This is the structure derived earlier, with equivalent kernel in eq. \ref{eq_BNN_per_RBF}. Since these BNNs output pure periodic functions they are unable to fit the data well.

\item \textbf{Combined BNN} - these models combined the ReLU \& Periodic BNNs from 1. and 2. above. For Mauna, an addition operation at outputs was applied, whilst for Airline, hidden nodes were multiplied point wise. Note that the main characteristics of the datasets are captured. \textbf{This creates sensible interpolation and extrapolation predictions.} Importantly, uncertainty increases with the time horizon.

\item \textbf{Combined GP} - the GPs corresponding to the combined BNNs in 3. were implemented. This enables verification that the BNN architectures produce a predictive distribution corresponding to the GP's (which could be thought of as the `ground truth'). The slight differences could be put down to the finite width of the BNNs, and imperfect inference procedure.

\end{enumerate}

\subsection{REINFORCEMENT LEARNING: PENDULUM SWING UP}

\begin{wrapfigure}{r}{0.3\linewidth}
\includegraphics[width=0.1\textwidth,scale=0.2]{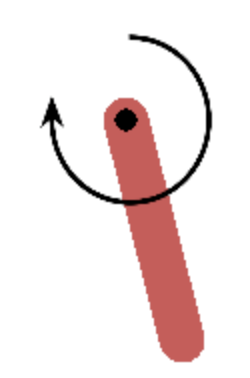}
\label{fig_pendulum1}
\end{wrapfigure}

We considered the pendulum swing up task; an agent applies torque to a bar on a pivot, maximising rewards by controlling the bar to be vertically upright. Observations consist of angle, $\theta$, and angular velocity, $\dot{\theta}$.

We used a slightly modified version of the task. Actions were discretised so that torque $\in \{-1,0,+1\}$. Dynamics were also modified - usually the update rule for $\theta$ is,
\begin{equation}
\theta_{t} = \theta_{t-1} + \dot{\theta}_t  dt,
\end{equation}
where $t$ is time, and $\dot{\theta}$ is a function of the applied torque and gravity. We modified this to,
\begin{equation}
\theta_{t} = \theta_{t-1} +  \frac{2}{1-e^{- \theta_{t-1}/3}}   \dot{\theta}_t  dt.
\end{equation}
This effectively introduces a frictional force that varies according to the absolute value of $\theta$. Crucially this means that as the bar spins, slightly different dynamics are experienced - this could arise from the bar spinning along a thread.

\begin{figure}[t!]
\begin{center}
    \begin{minipage}{0.5\textwidth}
        \centering
        \begin{overpic}[width=1.\textwidth]{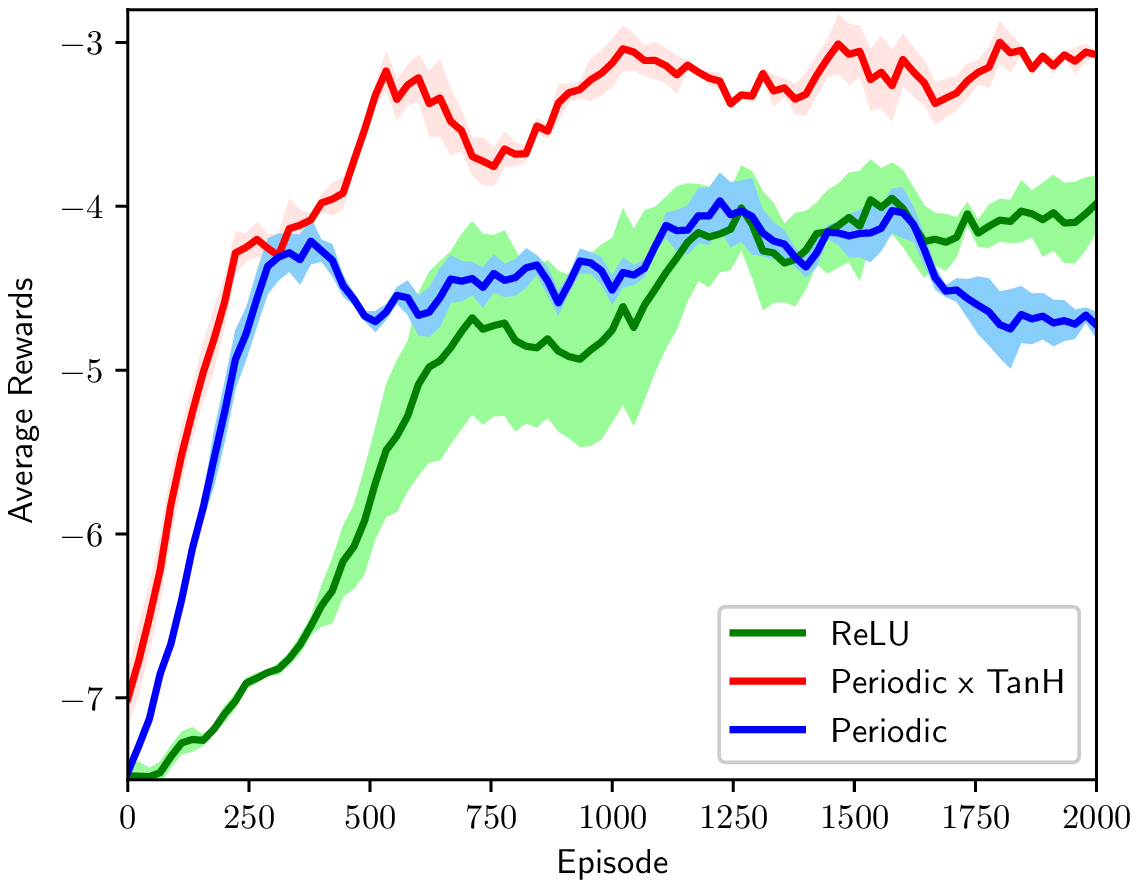}
        \end{overpic}
    \end{minipage}
\caption{Learning for three different BNN architectures on the pendulum task. The BNN incorporating a suitable prior for task, `Periodic$\times$TanH', outperforms basic BNNs. Mean $\pm$ 1 standard error, three runs.}
\label{fig_rl_train}
\end{center}
\vskip -0.2in
\end{figure}

A priori, we therefore know that the function is locally periodic. This makes the task challenging for basic BNN architectures - enforcing exact periodicity is just as inappropriate as ignoring it entirely.

We tested three BNN architectures on the task.
\begin{enumerate}
\item \textbf{ReLU}: a two-layer ReLU BNN with raw angle, $\theta$, and angular velocity, $\dot{\theta}$, as input. Priors: $\sigma^2_{w1}=\sigma^2_{b1}=1$, $\sigma^2_{w2}=\sigma^2_{b2}=1/50$, $\sigma^2_{w2}=\sigma^2_{b2}=10.0$.
\item \textbf{Periodic}: cos/sin input warping applied to $\theta$, raw angular velocity, $\dot{\theta}$, followed by a two-layer ReLU BNN. Prior variances as for 1.
\item \textbf{Periodic$\times$TanH}: takes the Periodic BNN as in 2., multiplied by a single-layer TanH BNN (taking only $\theta$ as input) with long length scale, $\sigma^2_{w1}=\sigma^2_{b1}=0.2$. This combines multiplication, warping and separation of inputs from section \ref{sec_kernel_combos_BNN}.
\end{enumerate}


Note that the benefits of BNN architecture should be independent of the learning algorithm and inference method. Here we used Bayesian Q-learning \citep{Dearden1998}, similar to regular Q-learning, but with Q-values modelled as \textit{distributions} rather than point estimates, with BNNs as the function approximators. 

It was important that a scalable technique be used for inference. Q-learning is sample inefficient, and the experience buffer accumulates hundreds of thousands of data points ($2,000$ episodes $\times$ $200$ time steps). Both GPs and HMC struggle with data of this size. We used Bayesian ensembles \citep{Pearce2018a, Pearce2018b} for inference - a recently proposed scalable, easily implementable technique.

Figure \ref{fig_rl_train} shows cumulative rewards for the three different architectures over 2,000 episodes. Periodic$\times$TanH clearly outperforms other models, both in terms of learning speed and quality of final policy. This is an example of the \textit{blessing of abstraction} at work - the more structure we account for, the less data we need \citep{Duvenaud2014} [p13]. The Periodic BNN has similar learning speed early on, but plateaus since it does not have the flexibility to fully capture system dynamics. ReLU, meanwhile, learns slowly, but has enough flexibility to capture closer to the true dynamics, and eventually surpasses the Periodic BNN. 

\begin{figure}[t!]
\begin{center}
	\hskip -0.15in
    \begin{minipage}{0.45\textwidth}
        \flushleft
        \begin{overpic}[width=1.05\textwidth]{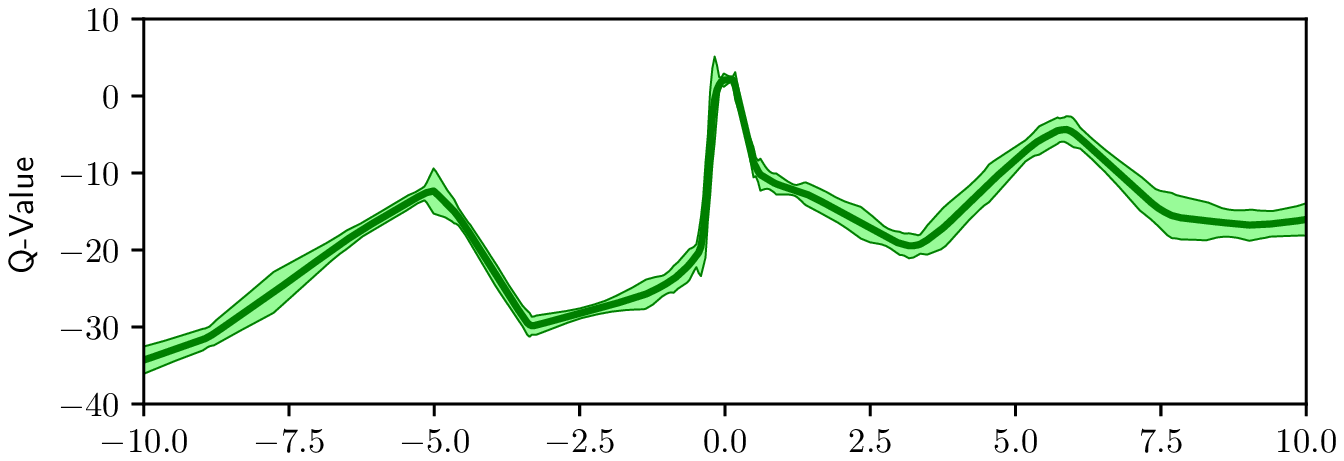}
         \put(82,8){ReLU}
        \end{overpic}
    \end{minipage}

    \vskip -0.05in    
    \hskip -0.15in
     \begin{minipage}{0.45\textwidth}
        \flushleft
        \begin{overpic}[width=1.05\textwidth]{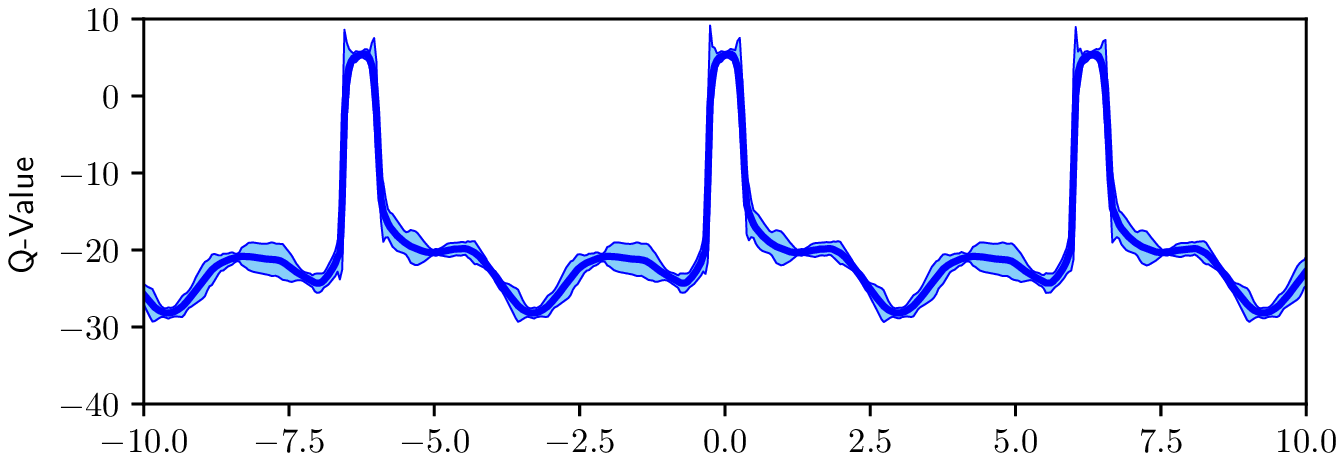}
         \put(79,8){Periodic}
        \end{overpic}
    \end{minipage}

    \vskip -0.05in
    \hskip -0.15in
     \begin{minipage}{0.45\textwidth}
        \flushleft
        \begin{overpic}[width=1.05\textwidth]{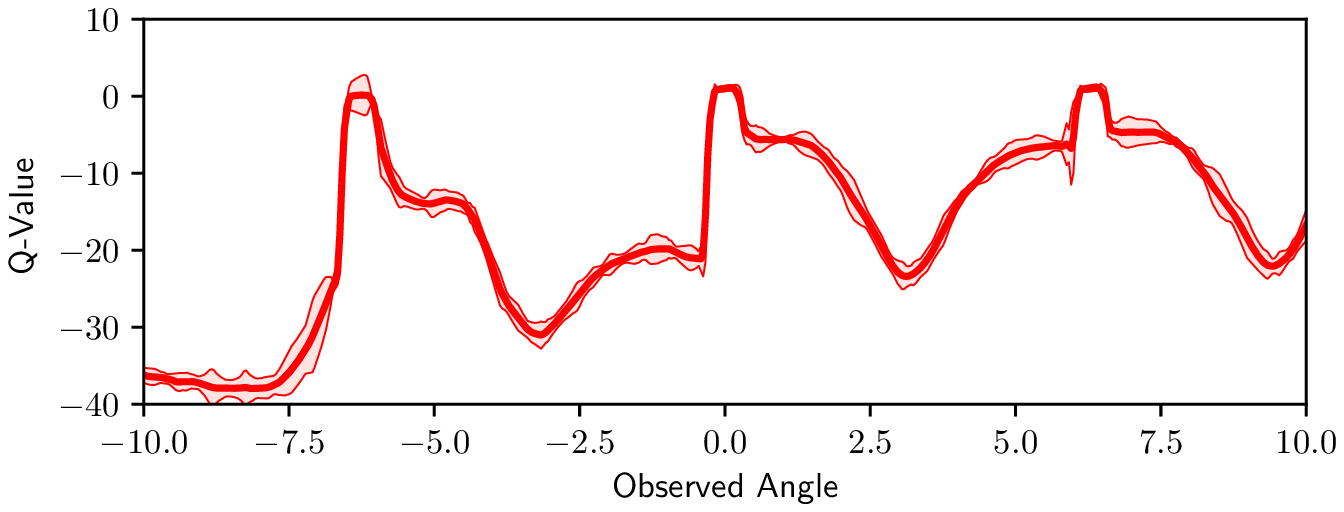}
         \put(65,11){Periodic$\times$TanH}
        \end{overpic}
    \end{minipage}
\caption{Q-values learnt for the action, torque $=0$, conditioned on observations of $\dot{\theta}= 0$, and input angle, $\theta$, varied on x-axis. Periodic$\times$TanH captures the local periodicity of the function.}
\label{fig_rl_surface}
\end{center}
\vskip -0.2in
\end{figure}

Figure \ref{fig_rl_surface} provides evidence for these comments. It shows the dynamics learnt for three revolutions of the pendulum for each BNN. The Periodic and ReLU BNNs are only able to approximate the optimum dynamics found by Periodic$\times$TanH.

\section{RELATED WORK}
\label{sec_related_work}
Two recent works proposed methods to overcome the limited expressivity of BNN priors. \cite{Flam-Shepherd2017} trained a BNN to output GP priors before running inference on a task. \cite{sun2019} had a similar approach that did not require pretraining. 

Both methods operate roughly in a supervised learning fashion, training BNNs to match the output of some GP, on training data augmented with sampled data points. In contrast, our approach directly incorporates priors into the model structure.





Several other works are of relevance. \cite{Ma2019} propose variational implicit processes for BNNs. \cite{Gaier2019} could be interpreted as fixing a posterior over parameters, and using evolutionary search to find a BNN architecture producing suitable posterior functions.

An orthogonal line of work to ours considers how to improve the scalability of GPs over the default $\mathcal{O}(N^3)$, e.g. \citep{snelson2006}. There also exist other techniques for creating expressive priors in GPs, e.g. \citep{Wilson2013}.

\section{CONCLUSION}
\label{sec_conclusion}
Expressive priors can be created in GPs by combining basic kernels into a new kernel. Noting the equivalence between GPs and infinitely-wide BNNs, this paper ported the idea to BNNs, deriving architectures that mirror such kernel combinations. Furthermore, we advanced the modelling of periodic functions with BNNs, which are often useful in this context.

These ideas are of practical benefit when some property is known about a function a priori, and basic BNNs do not model this well. We showcased two scenarios for which this was the case; time series modelling, and a RL task involving a locally periodic function.

In many learning tasks, a function's properties may be unknown or difficult to interpret, e.g. how does one specify priors in an Atari game learning from pixels?  Impact of our ideas could be amplified by research into automation of BNN design \citep{Duvenaud2013, Steinruecken2019}, and into how priors could be specified at a more abstract level.



\subsubsection*{Acknowledgments}
Thanks to the anonymous reviewers for their helpful comments and suggestions. The lead author was funded through EPSRC (EP/N509620/1).

\normalsize
\bibliography{library} 
\bibliographystyle{apalike}


\end{document}